\documentclass{article}
\usepackage{arxiv}
\usepackage{moreverb,url}
\usepackage{tabularx}

\usepackage[utf8]{inputenc} % allow utf-8 input
\usepackage[T1]{fontenc}    % use 8-bit T1 fonts
\usepackage{url}            % simple URL typesetting
\usepackage{booktabs}       % professional-quality tables
\usepackage{amsfonts}       % blackboard math symbols
\usepackage{nicefrac}       % compact symbols for 1/2, etc.
\usepackage{lipsum}
\usepackage{xcolor}
\usepackage{graphics}
\usepackage{graphicx}
\usepackage{amsmath}
\usepackage{rotating}
\usepackage{pdflscape}
\usepackage{caption}
\usepackage{subcaption}
\usepackage{makecell}
\usepackage{bbm}
\usepackage{booktabs}
\usepackage[english]{babel}

\usepackage[breaklinks,colorlinks,bookmarksopen,bookmarksnumbered,citecolor=red,urlcolor=red]{hyperref}

\begin{document}
\bibliographystyle{unsrt}

%\runninghead{Murray, Mitchell, Tuke, and Mackay}
%\runninghead{Author withheld}

\title{Revealing Patient-Reported Experiences in Healthcare from Social Media using the DAPMAV Framework}

\author{
  Curtis Murray \\
  The University of Adelaide\\
  \texttt{curtis.murray@adelaide.edu.au} \\
  %% examples of more authors
   \And
 Lewis Mitchell \\
  The University of Adelaide\\
  \texttt{lewis.mitchell@adelaide.edu.au} \\
  \And
 Jonathan Tuke \\
  The University of Adelaide\\
  \texttt{simon.tuke@adelaide.edu.au} \\
  \And
 Mark Mackay \\
 James Cook University\\
  \texttt{mark.mackay@unisa.edu.au} \\
}

%\author{Author withheld for anonymized peer-review.\affilnum{1}}
%\affiliation{\affilnum{1}Withheld.}
%\corrauth{Withheld}
\maketitle
\begin{abstract}
  Understanding patient experience in healthcare is increasingly important and desired by medical professionals in a patient-centered care approach. Healthcare discourse on social media presents an opportunity to gain a unique perspective on patient-reported experiences, complementing traditional survey data. These social media reports often appear as first-hand accounts of patients' journeys through the healthcare system, whose details extend beyond the confines of structured surveys and at a far larger scale than focus groups. However, in contrast with the vast presence of patient-experience data on social media and the potential benefits the data offers, it attracts comparatively little research attention due to the technical proficiency required for text analysis. In this paper, we introduce the Design-Acquire-Process-Model-Analyse-Visualise (DAPMAV) framework to provide an overview of techniques and an approach to capture patient-reported experiences from social media data. We apply this framework in a case study on prostate cancer data from /r/ProstateCancer, demonstrate the framework's value in capturing specific aspects of patient concern (such as sexual dysfunction), provide an overview of the discourse, and show narrative and emotional progression through these stories. We anticipate this framework to apply to a wide variety of areas in healthcare, including capturing and differentiating experiences across minority groups, geographic boundaries, and types of illnesses.
\end{abstract}


\section{Introduction}

Health systems in countries like Australia represent a significant investment of public resources \cite{aihwexpend}. There has been increasing interest in not only the activity that such systems generate but also in better understanding and measuring patient experience and patient outcome \cite{arah2003conceptual, murray2000framework, anhang2014examining, manary2013patient}. Historically, there has been a focus on using mortality as a measure of outcome \cite{McDonald68}. If greater or fewer people die while in the care of a facility, the facility can estimate that it is performing worse or better respectively. Alone, however, this outcome provides an insufficient view of healthcare for complete performance monitoring, as it overlooks many other aspects of healthcare,  and potentially only captures ill performance once lives have already been lost \cite{lilford2010using, barker2012value}. To attain a more holistic approach to performance monitoring, the Australian Health Performance Framework (AHPF) identifies indicators of health system performance as accessibility, appropriateness, continuity of care, effectiveness, efficiency and sustainability, and safety \cite{ahpf}. As such, these indicators are incorporated into health systems performance monitoring and are effectively measured by hospitals. However, the perspective of a hospital alone is incomplete for a comprehensive assessment of the health system's performance; there are some aspects of care where patients are the best judges of their health-related outcomes \cite{nemeth2006health}.

\subsection*{Patient-reported experiences are useful}

 Areas of healthcare where patients may be the best judge include health-related quality of life, physical function, psychological well-being, subjective symptoms such as pain, social function, and cognitive function \cite{nemeth2006health}. The World Health Organisation has identified the importance of patient engagement in benefiting healthcare systems, processes, and policies \cite{world2017technical}. Patient-reported outcome measures (PROMs) and patient-reported experience measures (PREMs) have been developed to capture patient feedback and address the gap in healthcare performance monitoring by harnessing the patient's perspective \cite{prosandprems}. This feedback is vital in providing quality healthcare that matches the public's expectations. It facilitates the alignment of resources with patient priorities and can act as an accountability structure \cite{commonwealthofaustralia_2019}. Due to its importance, patient-reported feedback is used in performance indicators for many institutions \cite{NHIPCC2017}.
 
%   Patient-reported outcomes (PROs) detail the outcomes of care provided using direct accounts from patients. Patient-reported experiences focus on the perception of the patient's experience in the healthcare setting. These are not outcome-based. Instead, they capture patients' personal healthcare experiences from their perspectives. This experience is also distinct from patient-reported satisfaction, culminating in how experience and outcomes matched the patient's expectations.

\subsection*{Traditional methods of capturing patient-reported experience overlook insightful information}

Focus groups and surveys are the traditional tools used to capture patient experiences. Focus groups form the qualitative arm to explore patient experiences, that can be used to direct quantitative surveys \cite{masadeh2012focus}. This dualistic approach of exploring, and then quantifying, is symbiotic, as focus groups and surveys can cover many of each other's weaknesses.

The use of open-ended questions in focus groups gives patients the freedom to discuss relevant issues in depth, however, due to the hands-on nature, this method is costly, difficult to scale, and lacks generalisability to the broader population \cite{vicsek2010issues}. On the other hand, surveys scale readily, however, suffer from their own weaknesses. To take an example, the Australian Commission on Safety and Quality in Healthcare released the Australian Hospital Patient Experience Question Set (AHPEQS) -- a core set of satisfaction questions -- as a tool that hospitals can use to capture patient-reported experiences \cite{ACSQHC}. This survey captures aspects of patient feedback related to general satisfaction in a multiple-choice format. For example, Question 1 states `My views and concerns were listened to', with the prescribed responses `Always', `Mostly', `Sometimes', `Rarely', `Never', and `Didn't apply'. Other questions relate to needs being met, feeling cared for, feeling involved, feeling informed about care, communication, pain management, confidence in safety and care, unexpected harm, and overall quality of care, all with similar programmed answers. The full set of questions is provided in the \nameref{ap:sup} in Table \ref{tab:survey}.

While surveys such as the AHPEQS capture specific areas of patient feedback, they are often restricted by their predefined scope and generality. These rigid surveys can result in blind spots in understanding the patient journey. For example, the AHPEQS asks whether `individual needs were met', but there is no scope to capture the patient's specific needs. The AHPEQS also asks how often the patient `felt confident in the safety of their treatment and care', yet does not facilitate the patient to outline when or where they felt unsafe. 

Deployment of these surveys is time-consuming, involving research and approval to ensure that the survey meets specific needs. While this is to the benefit of a survey in its ability to capture necessary information, it leads to a delayed deployment. Hence the survey may miss critical, time-dependent data. To ensure that patient-reported experiences are captured more thoroughly through surveys, sufficiently long surveys with well-poised questions are necessary. Survey length, however, is negatively associated with respondent rate \cite{fan2010factors}; a balance must be struck between respondent rate and survey length. 

Recently, surveys have begun to blend the quantitative with the qualitative, by including open-ended questions on patient experience. In this way, surveyors can aim to simultaneously receive quantitative data on patient experience, whilst also driving the qualitative exploration needed to overcome the would-be blind spots. However, open-ended questions introduce an additional layer of difficulty to patients, and as a result, can be overlooked.

Both focus groups and surveys suffer from common weaknesses. Raw data, and unpublished results, are often obscured due to ethics considerations. Whilst this is critical to patient confidentiality, it results in a fractured and occluded view of patient experience. Consequently, researchers interested in specific yet less explored areas of healthcare and patient experience may rely on interpretations derived from others' preliminary studies. This reliance on interpretations poses challenges and may require additional costly efforts, such as conducting their own focus groups or surveys, to gather data directly related to their research interests.

In order to effectively capture a more holistic view of patient experience, patients need the unconstrained freedom to discuss relevant to them, at an appropriate scale, with accessibility to other researchers. Currently, neither surveys nor focus groups, alone or combined meet this need.

\subsection*{Social media is a rich source of information}

Social media hosts vibrant forums featuring the discussion of topics related to social issues \cite{doi:10.4258/hir.2015.21.2.67, rainie2012social}. These discussions are information-rich and may provide meaningful insights into how complex issues are perceived at individual and community levels. As users determine discussion topics and their details, the discussion often corresponds to social issues relevant to their respective communities \cite{bowen2017advocacy}. The broad coverage of issues is directly aligned with areas of social concern. As many of these insights have the potential to be highly valuable, there is a growing demand for their capture \cite{ZHOU2018139,pappa2019harnessing,greaves2013harnessing}. Social media is, therefore, rapidly being adopted by researchers as a tool to explore these social issues to uncover valuable insights into public opinion \cite{anstead2015social,dodds2011temporal, murray2020symptom}. 
However, social media data is unstructured. This lack of structure introduces a challenge, as structured data is far easier to analyse at scale. Technological improvements in natural language processing aid in revealing latent structure in otherwise unstructured data. While this means researchers have pathways to analyse social media and other unstructured data at scale, this pathway is often inaccessible to those without a technical background. There is a need for guidance in analysing unstructured data for researchers with a non-technical background.

\subsection*{Social media hosts patient-reported experiences}

Discussion around healthcare is widespread on social media platforms such as Twitter, Facebook, and Reddit \cite{househ2014empowering, greaves2013harnessing}. Here, patients have the scope to discuss what is important to them in as much detail as they want. Online communities offer a sense of social inclusion through shared experience, that can often be sought as a result of suffering and social marginalisation \cite{davison2000talks}. As a result, discussion in these communities is of distinct value to understating patient experiences. Due to its anonymity and lack of direct face-to-face interactions, social media has been evidenced to elicit high levels of self-disclosure in people, who tend to reveal themselves more intimately than in face-to-face interactions \cite{joinson2007self}. This intimacy through community and anonymity, as well as public accessibility, makes social media a rich and open source of information for revealing patient-reported experiences to researchers. Healthcare systems may utilise this information to improve their facilities and procedures. Researchers have called for the mining of social media data as a complementary tool to improve healthcare systems \cite{greaves2013harnessing}. To aid future researchers in deciding which platform to gather patient experience data from, we summarise the research use of Twitter, Facebook, and Reddit and assess their appropriateness in the context of patient-reported experiences.

\subsection*{Social media research in patient-reported experiences}

Table \ref{tab:fb_tw_re} shows that there are over 15,000 Google Scholar search results for articles related to \emph{patient experience}. Examples of these can be seen for Facebook \cite{a2021assessing}, Reddit \cite{okon2020natural, du2020leaked, murray2020symptom}, and Twitter \cite{hawkins2016measuring, bovonratwet2021natural, clark2018sentiment}.

\begin{figure}
    \centering
    \includegraphics[width=0.5\textwidth]{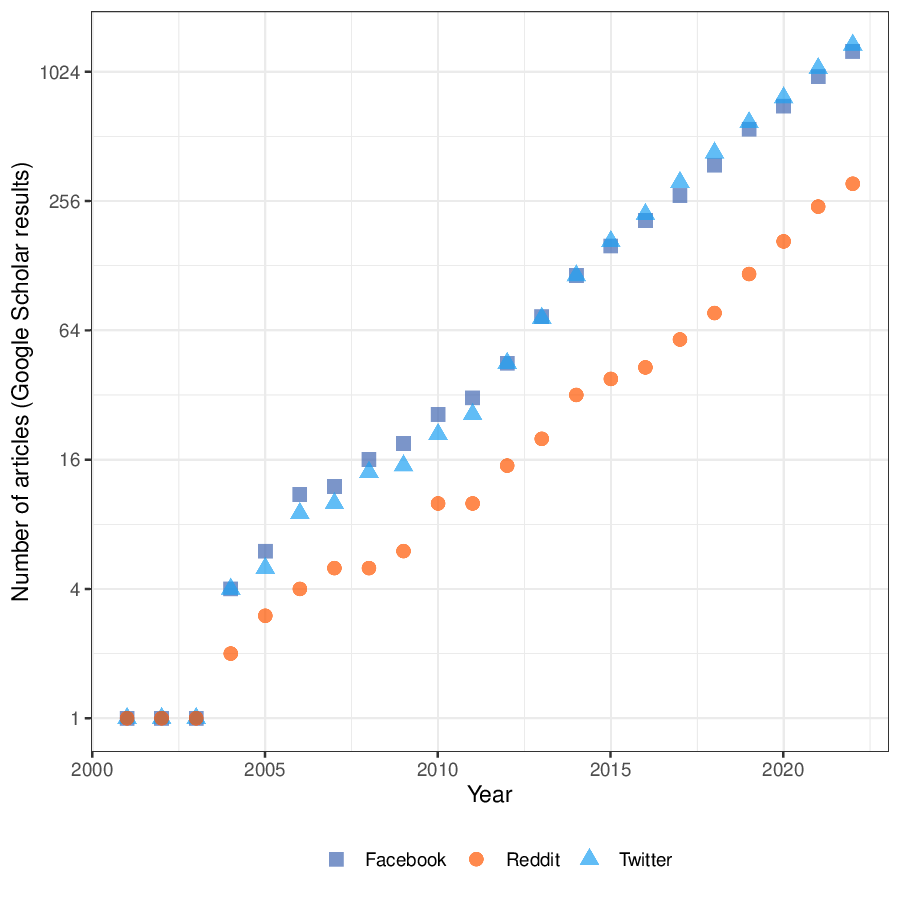}
    \caption{Cumulative Google Scholar search results for articles related to \emph{Natural Language Processing} and \emph{patient experience} for social media platforms; Facebook, Reddit, and Twitter. A binary logarithm transformation has been applied to the y-axis.}
    \label{fig:apy}
\end{figure}

There is a rapid growth of social media research related to patient-reported experiences. Figure \ref{fig:apy} shows exponential growth in the number of Google Scholar search results for a combined search of \emph{Natural Language Processing} and \emph{patient experience}. These Google Scholar searches were made using the following search terms under conditions that will be subsequently described.

\begin{verbatim}
   [social media platform] AND `natural language processing' OR `NLP' AND 
   `patient experience' OR `patient experiences' OR `patient reported 
   experience' OR `patient reported experiences'
\end{verbatim} 

The text \texttt{[social media platform]} was replaced with `Facebook', `Reddit', and `Twitter'. The search was filtered by date to find the cumulative number of articles that had been published by each successive year.

\subsection*{Social media research overview}

\begin{table*}
\caption{Number of daily active users \cite{facebookusers, redditusers, twitterusers} and Google Scholar search results for Facebook, Reddit, and Twitter using search terms \emph{patient experience} and or \emph{Natural Language Processing}. Retrieved 29 June, 2022. Table \ref{tab:key} shows the exact Google Scholar search terms, corresponding to the column \emph{Search Key}}
\label{tab:fb_tw_re}
\centering
\begin{tabular}[t]{>{\raggedright\arraybackslash}lp{5cm}lll}
\toprule
& Search Key (See Table \ref{tab:key}) & Facebook & Reddit & Twitter  \\
 \midrule
\bf{Daily Active Users} & - & 1,920,000,000 & 50,000,000 & 229,000,000\\
\midrule                                                 
 & - & 6,790,000 & 2,150,000 & 7,580,000  \\
 & Patient Experience & 17,400 & 15,200 & 17,700 \\
 \bf{Google Scholar} & Natural Language Processing & 82,300 & 20,900 & 119,000 \\
 \addlinespace
 & Patient Experience + Natural Language Processing & 1,050 & 260 & 1,130  \\
\bottomrule
\end{tabular}
\end{table*}

\textbf{Twitter}, a popular social media platform, is often the subject of research \cite{karami2020twitter}. On Twitter, users post online in the form of \textbf{tweets}. Prior to October 2018, these tweets were limited to 140 characters and now are limited to 280 characters. Tweets may contain \textbf{hashtags} that allow users to tag keyword phrases that correspond to specific topics or issues. \textbf{Reddit}, colloquially known as the `front page of the internet', is another social media platform that receives attention from researchers \cite{medvedev2017anatomy}. Discussion on Reddit occurs in a different form from that on Twitter. On Reddit, users post to communities that cover specific topics, known as \textbf{subreddits}. A user may create a new post to a subreddit or may post a reply to another user's post or reply. \textbf{Facebook} is another social media platform that attracts attention from researchers. On Facebook, discussion often occurs in the form of posts and replies made to users' walls, groups, businesses' walls, and other pages. 

Twitter is the most frequent subject of research, with approximately 1.1 times and 3.5 times more Google Scholar search results than Facebook and Reddit respectively, as seen in Table \ref{tab:fb_tw_re}. One possible mechanism for Twitter's research popularity over Reddit's could be Twitter's greater popularity, having 4.49 times more daily users than Reddit, also seen in Table \ref{tab:fb_tw_re}. With almost ten times more users than Twitter, Facebook is underrepresented in research articles per daily active user. This underrepresentation may be explained by restrictions imposed on Facebook's data accessibility in response to the 2018 Cambridge Analytica data scandal \cite{bruns2018facebook, schroepfer2018update}. The widespread use and data policies of Twitter and Reddit make them suitable platforms for research over other platforms such as Facebook, whose data policies are more prohibitive for research \cite{lockdown2019}.

Twitter has no prescribed community structure for the discussion of specific topics. Topics on Twitter are optionally discussed through the use of hashtags. This approach allows posts to be identified as belonging to multiple topics simultaneously. Reddit, contrastingly, has no obvious method for cross-topic discussion. Reddit's structure instead necessitates posting to exactly one subreddit. This subreddit usually hosts most of the discussion around that topic, creating a sense of community. One study noticed that in the context of the `Me too' movement, discussion on Twitter was less personal than on Reddit and instead focused on encouraging others to speak up \cite{manikonda2018twitter}. Reddit's lack of post character limit that Twitter enforces may make it a more suitable platform for deeper discussion surrounding healthcare, as users do not need to restrict the content they share to fall under the character limit.

\subsection*{Challenges of using social media data}

Surveys offer a highly structured medium for collecting public opinion data. This structure arises from targeted questioning coupled with predefined multiple-choice answers, allowing responses to be easily tallied and analyzed. However, free-text data, such as those found in social media, lack this structure, posing a significant challenge for eliciting patient-reported experiences \cite{tanwar2015unravelling}. While focus groups can manually overcome the challenge of unstructured responses due to their smaller scale, this approach quickly becomes infeasible as the volume of data increases. Natural Language Processing (NLP) models aim to address this lack of structure in large free-text datasets. By employing statistical and machine-learning models, NLP uncovers latent structures within the data, revealing patterns or trends across corpora that provide an 'at-a-glance' understanding of the content, eliminating the need for close reading \cite{blei2003latent}. Once this latent structure has been uncovered, the data becomes ripe for analysis.

\subsection*{The need for a guiding framework}

Natural language processing tools are often applied by researchers with a technical background in computer science or mathematics. This creates a barrier for many healthcare researchers and practitioners who, while possessing the domain knowledge to derive valuable insights from these tools by mining\footnote{Analysing and drawing insights from data at scale.} social media texts to uncover patient-reported experiences, may be unaware of or lack the technical background to use the tools effectively.  This dichotomy motivates the need for a generalised reproducible framework that empowers healthcare researchers to exploit free-text narratives. We introduce the Design-Acquire-Preprocess-Model-Analyse-Visualise (DAPMAV) framework to address this need. The DAPMAV framework provides an overview of, and a structured approach to mining patient-reported experiences from social media discourse, acting as a qualitative exploration that is complementary to surveys. 

We apply this framework to a discussion on prostate cancer and demonstrate that the insights found through the qualitative analysis correspond to known areas where patient-reported experiences are sought. Aside from non-melanoma skin cancers, prostate cancer is Australia's most commonly diagnosed cancer (21,808 diagnoses in 2009). It is also the fourth leading cause of death amongst Australian men (3,294 deaths in 2011) \cite{australian2013prostate}.

\begin{figure*}
 \centering
 \includegraphics[width=0.8\textwidth]{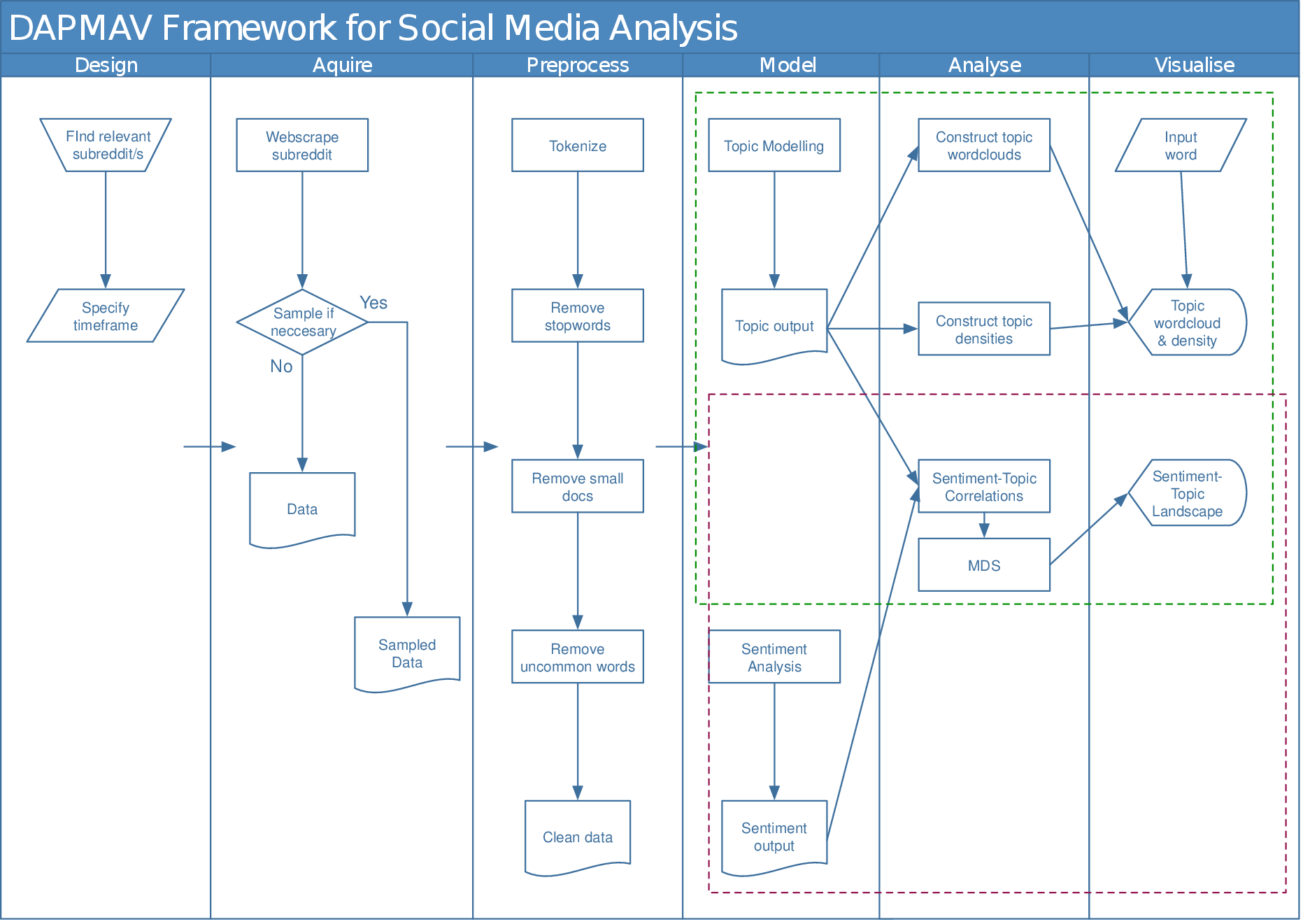}
 \caption{Design-Acquire-Process-Model-Analyse-Visualise (DAPMAV) framework for analysis of social media discussion relating to patient-reported experiences.}
  \label{fig:swim}
\end{figure*}

\section{DAPMAV Framework}

\subsection*{D - Design}

The design stage is the first stage in the DAPMAV process, summarised in the first column of Figure \ref{fig:swim}. In the design stage, we choose a relevant social media data source. In this example, we will walk through the steps to find relevant data on Reddit. Following this choice, we select the time frame to collect the posts in, and how to handle metadata.

\begin{figure*}
\centering
    \begin{subfigure}{0.7\linewidth}
 \includegraphics[width=\textwidth]{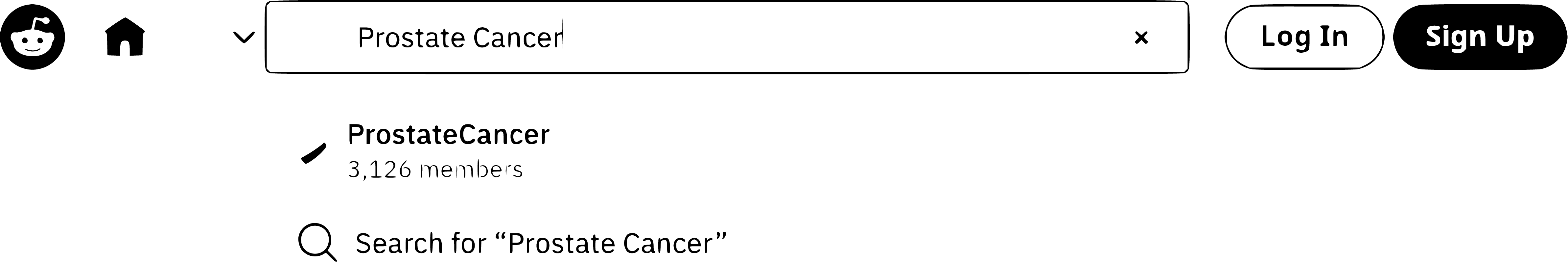}
 \caption{Using the Reddit search bar to search for Prostate Cancer. Once the search term is entered, we can click \emph{Search for `Prostate Cancer'} to perform the search. Results are seen in Figure \ref{fig:reddit_search2}.}
  \label{fig:reddit_search}
    \end{subfigure}
\hfil
    \begin{subfigure}{0.7\linewidth}
 \includegraphics[width=\textwidth]{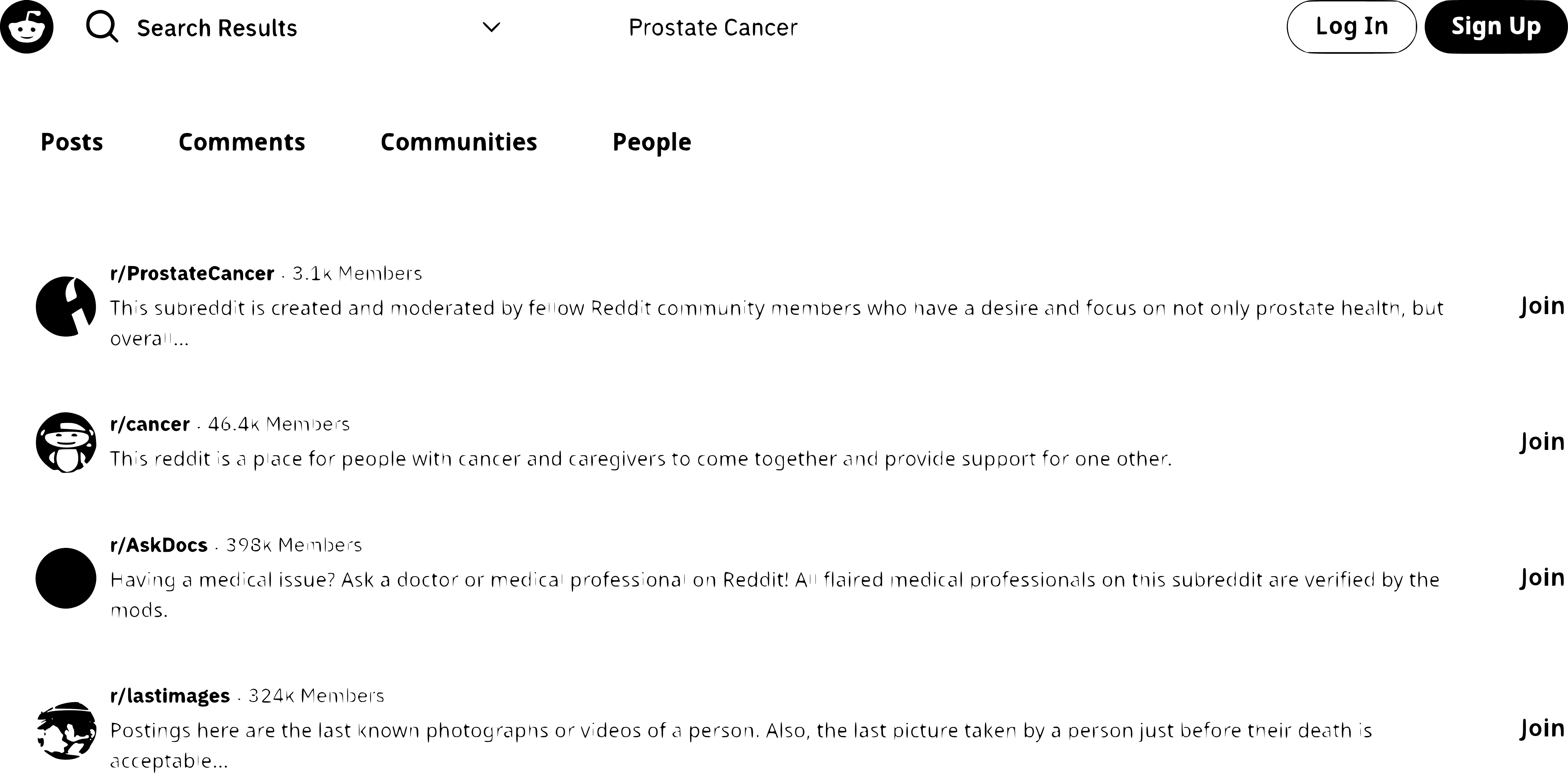}
 \caption{Reddit search results for \emph{Prostate Cancer}, performed in \ref{fig:reddit_search}. The first result is a community titled /r/ProstateCancer. We click on this entry to navigate to the /r/ProstateCancer subreddit, seen in Figure \ref{fig:reddit_pc}.}
  \label{fig:reddit_search2}
        \end{subfigure}
    \par\bigskip \par\bigskip
        \begin{subfigure}{0.7\linewidth}
 \includegraphics[width=\textwidth]{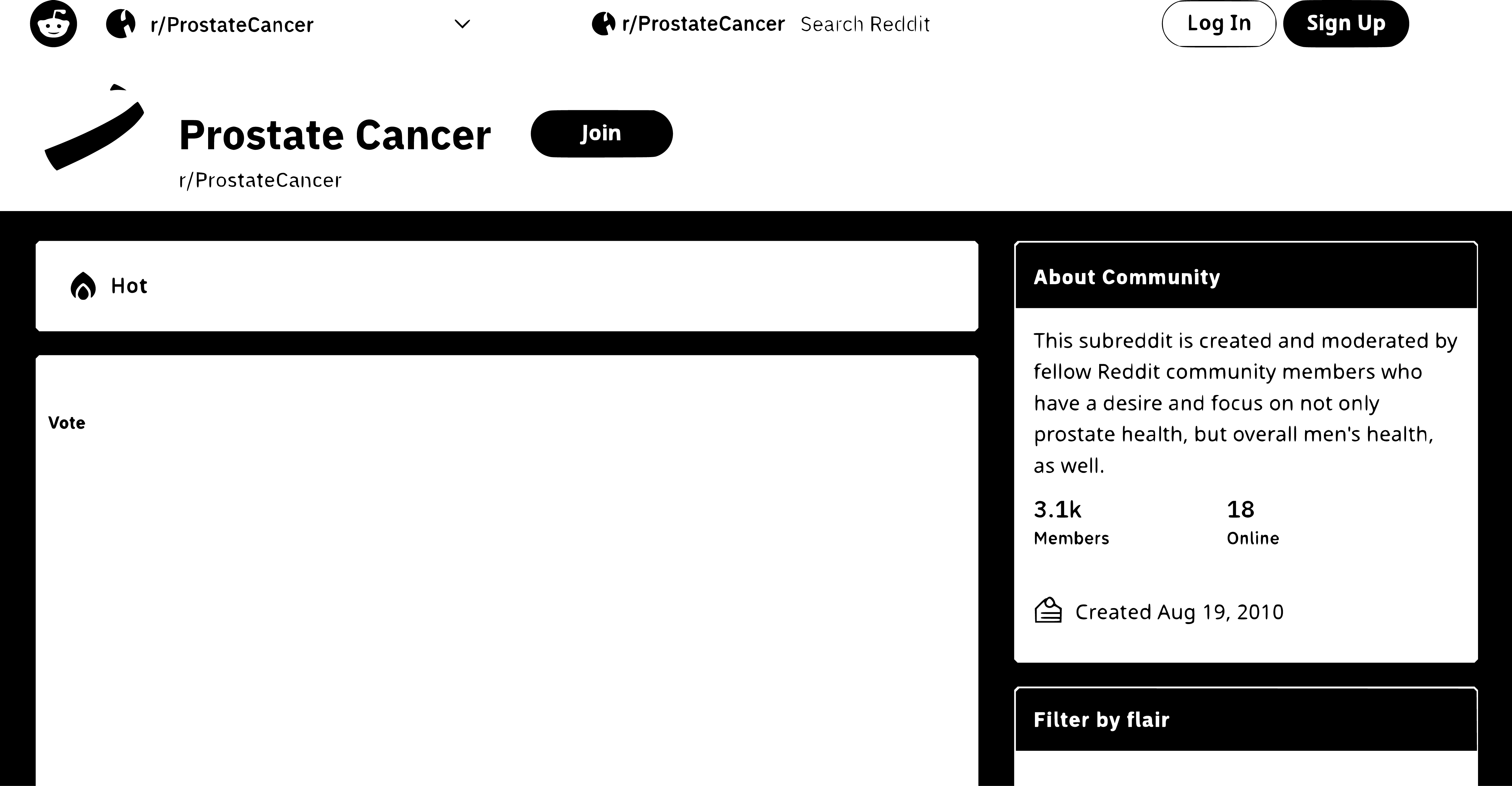}
 \caption{Prostate Cancer subreddit resulting from the searches made in Figures \ref{fig:reddit_search}-\ref{fig:reddit_search2}. Posts can be seen in the left column (blurred for anonymity), and details about the community can be seen in the right column.}
  \label{fig:reddit_pc}
    \end{subfigure}
    \caption{Process of finding a relevant subreddit.}
        \label{fig:reddit_all}
    \end{figure*}

One way to select the subreddit would be to use the Reddit search bar to search for the overall theme of the discussion and select a relevant community from the communities tab. Consider for this example that we are interested in observing prostate cancer discussion. We type `Prostate Cancer' into the search bar. This is seen in Figure \ref{fig:reddit_search}. The results of this search are seen in Figure \ref{fig:reddit_search2}, where we have navigated to the communities tab. We notice a specific prostate cancer community called `/r/ProstateCancer'. We can click on this community to be taken to the subreddit (Figure \ref{fig:reddit_pc}). We see posts made by Reddit users, an \emph{About Community}, as well as \emph{flairs}, and post tags entered by users that can be filtered. We can explore the posts to ensure they are relevant to the conversation we want to capture. If so, we move on to specifying our timeline.

The \emph{About Community} box on the right-hand side of Figure \ref{fig:reddit_pc} shows that the subreddit was created on August 19, 2010. This date may be important as some subreddits may be too new for their intended research purposes.

Posts on Reddit can feature chains of comments, known as threads. Any comment can itself be commented on, forming a comment tree for each post. This leaves researchers with the design choice of how to handle these threads. In large communities with many posts, it may be sufficient to only take the original posts, and omit comments. In other cases, where there are few posts to the community, capturing discussion threads may be beneficial and bolster the number of documents that are retrieved for analysis. 

\subsection*{A - Acquire}

Data acquisition is the second stage of the DAPMAV process, summarised in the second column of Figure \ref{fig:swim}. In the data acquisition process, we retrieve the Reddit posts in the subreddit that we wish to analyse from the period specified in the design phase. We obtain the Reddit posts using the Pushshift Application Protocol Interface (API), an interface to the Pushshift archiving platform that provides researchers access to Reddit data \cite{baumgartner2020pushshift}. The choice of the Pushshift API may be favourable to Reddit's API as it offers a greater rate of post-acquisition \cite{baumgartner2020pushshift}.

Social media datasets are vast and may have entries in the order of millions or billions. Computationally demanding algorithms inhibit the rapid exploration of large text datasets. To avoid lengthy processing, we can thin out our text data to reduce the size of the dataset, if necessary. As a means of thinning out our dataset, we can sample a fraction of the documents in our corpus without replacement. By sampling in this fashion, we obtain a representative subset of documents across our timeline with reduced computational overhead. The number of samples we take will determine the speed of our analysis. The amount that is thinned out, if thinning is necessary at all, will be determined by the researcher's intent. If, in the later analysis, we find results lack specificity, we may want to thin the dataset out less. For example, if a community features hundreds of thousands of posts, we may wish to sample a few thousand posts for an initial rapid analysis, decide if we wish to continue using this dataset, and if so, potentially increase the sample size to tens of thousands for a more in-depth analysis.

Reddit posts also may be tagged in certain ways. As an example, the subreddit `/r/COVID19Positive' can have posts tagged with the flairs `Question-for medical research', `Tested Positive - Family', `Presumed Positive - From Doctor', `Tested Positive - Me', `Question-to those who tested positive', and finally `Tested Positive'. These flairs give us additional data about posts we may want to consider. In particular, if we are only interested in posts relating to a particular flair, we can filter out the other flairs to reduce the run time of the models and potentially improve the specificity of the results. However, in cases with limited data, including additional yet largely relevant flairs may improve the model's performance.

\subsection*{P - Preprocess}

Preprocessing is a preliminary form of data processing prior to the principal analysis that prepares the data to be in a form that is suitable, effective, and efficient for the required analysis \cite{garcia2015data}. Efficiency is primarily centred around finding a representation of the data that reduces the feature space while maintaining as much relevant information as possible. In this section, we outline basic procedures typically used for text analysis \cite{silge2017text, gerlach2018network}. We make use of the programming language R \cite{ihaka1996r}, as well as several packages from the Tidyverse \cite{wickham2019welcome} and TidyText \cite{silge2016tidytext} libraries. Our preprocessing code is available in the GitHub repository \cite{murray2022DAPMAV_github}.

\subsubsection*{Tokenise}

\textbf{Tokenisation} is the process of deconstructing text into a list of distinct units, called \textbf{tokens} \cite{silge2017text}. A token may be a unigram (a single word), a bi-gram (two consecutive words), more generally, an n-gram (n consecutive words), a sentence, and many other things.

\subsubsection*{Remove Stop words}

When conducting text analysis, we may wish to remove a list of words, denoted \textbf{stop words}, from the corpus \cite{blei2003latent}. Words that are often considered stop words are common words that add little contextual information. Examples of stop words include `the', `and', `of', and `in'. Removal of these words decreases the dimensionality of the data, thereby reducing the computational complexity and hence the time taken to model.

\subsubsection*{Remove Small Documents}

Excessively small documents may be removed in text preprocessing \cite{WANG2018142}. Small documents may not contain sufficient information to be valuable, depending on the context of the data and the research question. Including small documents may dilute the effectiveness of the results while also increasing the computational complexity. Researchers will benefit from assessing their needs based on the context of their problem and the size of their dataset when deciding whether or not to set a minimum document length threshold.

\subsubsection*{Remove Uncommon Words}

Removal of uncommon words is another common preprocessing technique to reduce the feature space \cite{nguyen2015improving}. Word frequencies typically follow a Zipfian distribution, and are heavily tailed, with a large number of rare words \cite{piantadosi2014zipf}. As a result, removing only exceedingly rare words (for example, those with less than three occurrences) can significantly reduce the feature space without compromising critical information.

\subsection*{M - Model}

\subsubsection*{Topic Modelling}

\begin{figure*}
    \centering
    \includegraphics[width=0.8\textwidth]{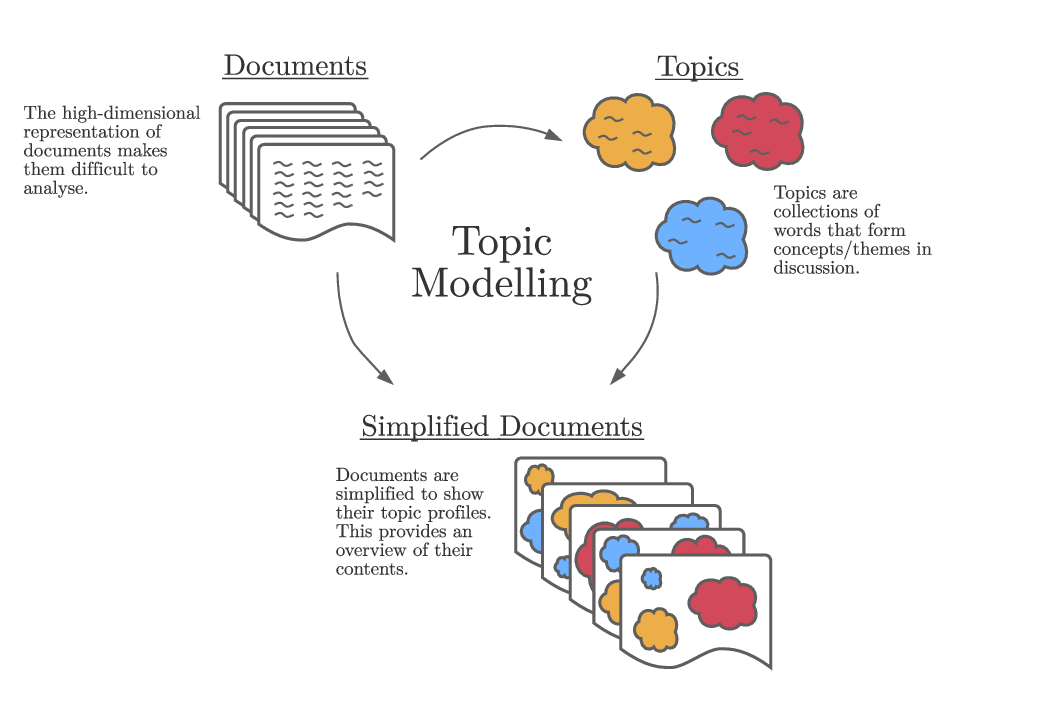}
    \caption{Topic Modelling Overview}
    \label{fig:tm_overview}
\end{figure*}
  
Large amounts of unstructured text data pose a challenge to researchers as their high complexity masks information \cite{tan1999text}. Tools in natural language processing (NLP) allow information to be uncovered and presented in a structured format that we can then analyse. Topic modelling is one such NLP tool that uncovers information in unstructured text data. Topic modelling can be viewed as a dimension reduction tool that projects high-dimensional text data onto low-dimensional spaces, peeling back complexity to reveal interpretable structure \cite{blei2003latent}. This dimension reduction occurs through the summarisation of text into topics or conversational themes. Documents (which can be social media posts) are reduced to mixtures of topics, providing an overview of their contents. Topics are represented by collections of words that are used in similar contexts. Figure \ref{fig:tm_overview} shows the topic modelling process, taking documents, finding appropriate topics, and then representing the documents as mixtures of these topics.

To illustrate the concept of topic modelling, consider a simplified example where we have a set of news articles. These news articles may generally cover news from three recurring themes that occur; the weather, finance, and politics. By performing topic modelling on this corpus, we may identify three topics, i.e. three sets of words, with each set corresponding to one of the news themes. In doing so, we can roughly measure the prevalence of each theme in any document.  For instance, an article discussing climate change may predominantly use words from the `weather' topic, with less use of words from the `politics' and `finance' topics. Conversely, an article about the housing market crisis would align strongly with the `finance' topic. Thus, topic modelling enables us to summarise the thematic composition of each document, while additionally allowing us to explore the semantic nuances of each identified topic.

\subsubsection*{Sentiment Analysis}

Sentiment analysis is another natural language processing tool that, similar to topic modelling, reduces the complexity of documents to a summary. However, instead of deconstructing documents into topics, sentiment analysis identifies sentiments and emotions conveyed in text. There are multiple ways to conduct sentiment analysis; for example, a simple yet effective approach is to use a predefined sentiment lexicon, such as AFINN \cite{nielsen2011new}, Bing \cite{hu2004mining}, or NRC \cite{mohammad2013crowdsourcing}, which prescribe sentiments, such as positivity and negativity -- but also potentially more complex emotions such as happiness, sadness, fear, anger, and more --, to words in a vocabulary. A document's sentiment can be found by aggregating the sentiments of each word in the document \cite{silge2016tidytext}. In this way, a document that uses a large degree of negative words compared to positive words will have a negative sentiment score. This is a simple, yet powerful technique, as the way a document is written by a patient, in regards to the sentiments of the words that are used, can reveal insights into the experiences of that patient \cite{greaves2013use}. Namely, it may help to show whether or not they had a positive or negative experience, and hence help to construct a more complete picture of their experience. When the sentiment scores of a corpus are viewed; either as a whole, or by making comparisons through time and strata, it can reveal insights into collective patient experience, and how it can vary in different ways.

\subsection*{A \& V - Analyse and Visualise}

In a grounded theory approach, where data drives the research, analysis and visualisation are heavily intertwined \cite{glaser2017discovery}. When dealing with large corpora, visualisation of the summaries found through topic modelling and sentiment analysis is a useful technique to qualitatively explore the structure of the narratives. Appropriate and effective visualisation can help to elucidate meaningful insights from the text. Observations made from the data through visualisation often evoke further questions. These questions can be used to cultivate hypotheses that may be answered through analysis. Results from the analysis may be explored through further visualisation, leading to more questions, hypotheses, and further analysis in a cyclical approach. As a result of this marriage between analysis and visualisation, we have elected to combine the analysis and the visualisation sections for more comprehensive descriptions that avoid repetition.

\subsubsection*{Analysing Topics and Sentiments}

Topic modelling can help to unravel the intricacies of social media healthcare posts. The levels of discussion for each topic tell us how often particular themes are discussed. Profiling documents in this way can yield multifaceted insights, encompassing overarching discussion themes, themes in individual documents, patterns relating to specific subgroups, longitudinal trends, and more. Similarly, sentiment analysis results can be analysed to see overall sentiment, sentiment in specific posts or groups of posts, and temporal changes in sentiment. Some useful approaches in topic and sentiment analysis are explained below.

\subsubsection*{Exploring Topic Contents} 

Exploration of the identified topics found through topic modelling serves as a foundational step in this analytical journey. By closely examining the topics that emerge from the discourse, researchers can establish connections between identified themes and preexisting domain knowledge. This process facilitates the assessment of the degree to which these topics align with or challenge conventional understanding. Notably, this exploratory phase enables the identification of insightful topics that may warrant further investigation, serving as starting points for more detailed analysis. 

Although topics can be presented in a precise fashion using tables of words with their densities, this method is visually cumbersome. Instead, topic word clouds can capture similar information in a simple visualisation. To produce a topic word cloud, we make use of the topic distributions found from topic modelling. These tell us the probability that a word is in a topic, $P( \text{word} | \text{topic})$. Visualisation of these probabilities through word clouds illustrates words contained in the topic. The size of a word in a word cloud relates to the probability that the word is selected from the topic. 

\subsubsection*{Codifying Topics into Themes} Beyond seeing topics as just lists of words (or word clouds), researchers can begin to carefully codify topics to capture the themes in the discussion that they represent. In doing so, researchers give topics contextual meaning by linking their rich details to well-understood concepts in their domain. This succinct representation fascilitates easier communication and interpretation of findings for further analyses.

\subsubsection*{Exploring Overall and Group Dynamics}

A multifaceted understanding of patient experiences can be found through a dual exploration of both overall discourse trends and specific group dynamics. By examining the prevalence of topics and sentiments within the entire corpus, and within specific subgroups, such as different demographics, regions, diseases, and more, researchers can identify dominant themes and emotions, and investigate variations that reflect the diversity of patient experiences.

Group differences in topic or sentiment densities can be observed through their log odds ratios. This quantitative approach enables researchers to compare the prominence of specific topics or sentiments and evaluate the relative strength of their variations between different segments of patients. Here, a log odds ratio near zero will show that there is little variation between groups, but a large negative or positive log odds ratio will indicate a strong difference between groups. These log odds ratios can be visualised, for example, by a bar chart. To assess for statistical significance in the difference, one could employ a Fischer test (or Chi-squared test) by constructing contingency tables for topic-word counts or sentiment-word counts within groups.\footnote{Topic-word counts and sentiment-word counts refer to the number of times words of particular topics or sentiments occur.}

\subsubsection*{Profiling Patient Experience}

Constructing patient profiles serves as a useful tool to help capture summaries of individual patient experiences. Assessing the prevalence and strength of certain thematic elements of a patient's narrative, and the sentiments it possesses, can help to build a unique profile that captures the individual's experience. Flagging patients as those that match elements of an interesting or concerning profile may allow researchers or healthcare practitioners to easily identify where they can conduct research, or provide an intervention strategy. For example, highly negative narratives, or those that make heavy use of a potential mental health topic, could be flagged for a human to assess if any intervention is necessary.

\subsubsection*{Unveiling Temporal Patterns}

Temporal topic and sentiment exploration delves deeper into the progression of discussions within the discourse and offers a dynamic perspective on the emergence and progression of narrative themes. This exploration may consider either the time in which posts were made, or each post's \textit{narrative time}, a time computed or derived from the narrative itself. For example, posts may be diarised, as seen in \cite{murray2020symptom}, allowing temporal information to be extracted. Alternatively, narrative time may be taken from the depth of a story using word positions, as was the case in \cite{reagan2016emotional}. By tracing how topics and sentiments evolve over distinct time intervals, researchers gain insights into the narrative arc that underpins patient experiences, helping to strengthen an understanding of how patient experience evolves through time. These results can be visualised through time-series plots, for example showing the change in sentiment through time, by constructing stacked histograms of topic-density through time.

\subsubsection*{Unravelling Topic Relationships}

Exploring the interconnectedness of topics illuminates the intricate relationships among thematic elements in the discourse. Researchers may employ multiple techniques to assess these relationships. Correlations between topics may be explored through quantitative methods, with closely correlated topics often emerging in the same documents. Alternatively, the network-based approach to topic modelling introduces a hierarchical stochastic block model (hSBM) that captures interactions between topics through a matrix of community-to-community interactions \cite{gerlach2018network,peixoto2014hierarchical,peixoto2017nonparametric}. This matrix captures the likelihood of edges between and within topic communities, unveiling the intricate relationships between topics within a structured network. By probing these relationships, researchers gain insights into themes that shape discourse patterns, which may help to reveal deeper insights into patient experience and the connections between different areas.

An effective way to quickly visualise these interactions is to create a topic map. This can be done by converting the interactions into a measure of dissimilarity and performing a dimension reduction of the high-dimensional topic interaction data down to two dimensions to be plotted. Such a map illustrates thematic relationships through spatial distances between topics, revealing clusters and patterns indicative of the underlying thematic structure. By offering a consolidated view of collective patient experience, this representation can help to communicate the core themes of patient experience to practitioners, researchers, and stakeholders. Traditional dimension reduction approaches such as principal component analysis (PCA) \cite{wold1987principal} and multidimensional scaling (MDS) \cite{kruskal1978multidimensional} attempt to preserve the pairwise distance structure of the high-dimensional data in a lower-dimensional projection. Modern approaches such as t-distributed stochastic neighbour embedding (t-SNE) \cite{van2008visualizing} and uniform manifold approximation (UMAP) \cite{mcinnes2018umap}, which focus on preserving local distances, not global distances, typically perform well, especially when viewing local structures. These approaches may not be as appropriate as MDS when the importance is on considering the global structure \cite{cox2008multidimensional}. 

\section{/r/ProstateCancer Data}

\begin{figure*}
  \centering
  \begin{subfigure}{0.48\textwidth}
    \centering
    \includegraphics[width=\textwidth]{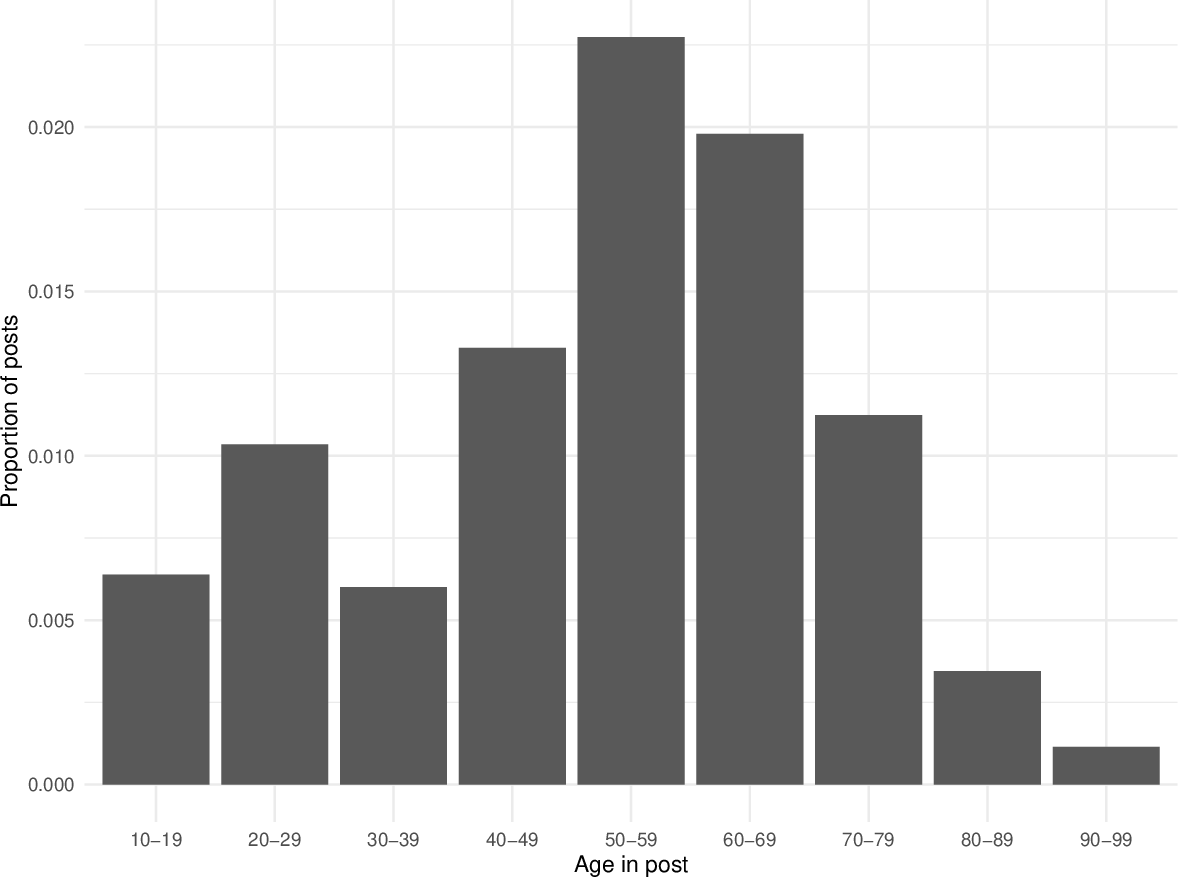}
    \caption{Age Profiles: Proportions of Age Range Mentions in /r/ProstateCancer.}
    \label{fig:ages}
  \end{subfigure}
    \hfil
  \begin{subfigure}{0.48\textwidth}
    \centering
    \includegraphics[width=\textwidth]{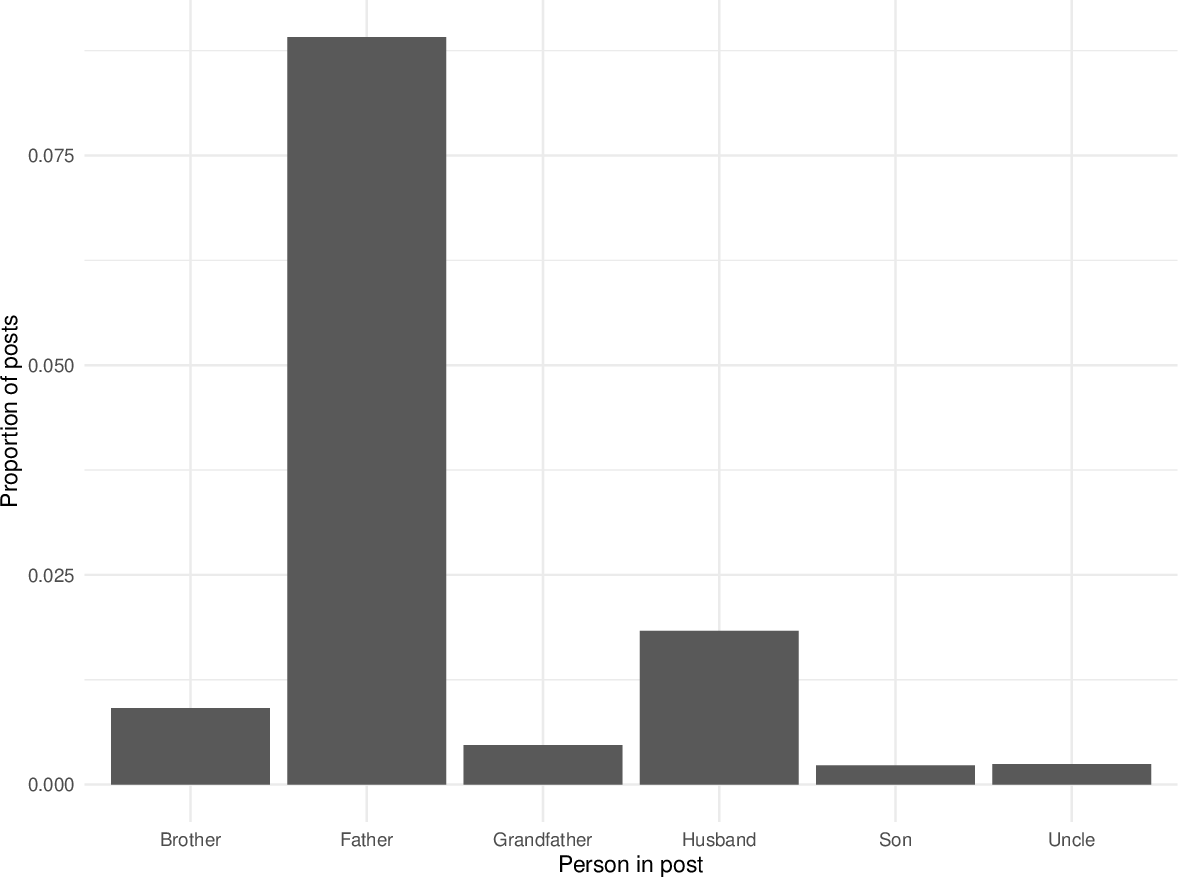}
    \caption{Subject Profiles: Proportions of Male Personas mentions in /r/ProstateCancer.}
    \label{fig:persons}
  \end{subfigure}
  \caption{Distribution of Age and Male Subject Mentions in /r/ProstateCancer.}
  \label{fig:age_and_subjects}
\end{figure*}

We follow the DAPMAV process to acquire $4074$ posts and replies from the Reddit community /r/ProstateCancer between 2019 and 2021. These consist of $631$ posts and their $3443$ replies. The mean number of non-stop words in the posts is $50.6$, and $31.2$ for their replies ($34.2$ combined mean). The longest post consists of 738 non-stop words. We restricted posts and replies to consist of at minimum ten non-stop words to remove documents with few words, as inspection showed documents with fewer than 10 non-stop words to have largely uninformative discussions. The prostate cancer stories posted to Reddit often feature mentions of the subject's age. String detection using Regular Expressions (REGEX) was used to estimate subjects' ages (in decade brackets). Figure \ref{fig:ages} illustrates the results of this. In addition to age, subjects of the narrative are identified by male family member names, seen in Figure \ref{fig:persons} REGEX was used to estimate the identity of the male family members: brother, father, grandfather, husband, son, and uncle.

% \begin{figure}[h!] 
%  \centering
%  \includegraphics[width=0.5\textwidth]{global.eps}
%  \caption{}
%   \label{fig:global}
% \end{figure}

% A word cloud is an effective tool for visualising word frequencies. Figure \ref{fig:global} is a word cloud that captures the overall conversation on /r/ProstateCancer, with the size of the words given as their empirical frequency.

\section{Results}

\subsection*{Topic Modelling}
Global word frequencies from a discussion give a broad overview of the discourse. While there is something to be gained from these word frequencies alone, they often struggle to capture the full context of the discussion they are involved in. Topic modelling captures this context by forming communities of words that are frequently used in the same context. We identify topics from the prostate cancer discourse that illustrate several revealing aspects of the discussion. These topics are manually summarised into a one- to two-word summary of what the topic appears to represent. Table \ref{tab:topics} shows this summary, the corresponding figure reference number, and the density of the topic.

Network topic modelling reveals the hierarchical structure of topics discussed in the prostate cancer data. In the supplementary materials, an interactive application displays this structure in a radial tree, where each root node is a word whose size is given by its empirical use, and hierarchical topic membership is defined by connections to parent nodes \cite{murraynetwork2022}. This topic structure allows for a detailed view of the discussion by breaking down the conversation into distinct topics that cover a particular aspect of the conversation. The following subsections will contain results relating to select topics found through the discourse and relate them to aspects of healthcare. 

\begin{figure*}
\centering
    \begin{subfigure}{0.49\linewidth}
        \includegraphics[width=\linewidth]{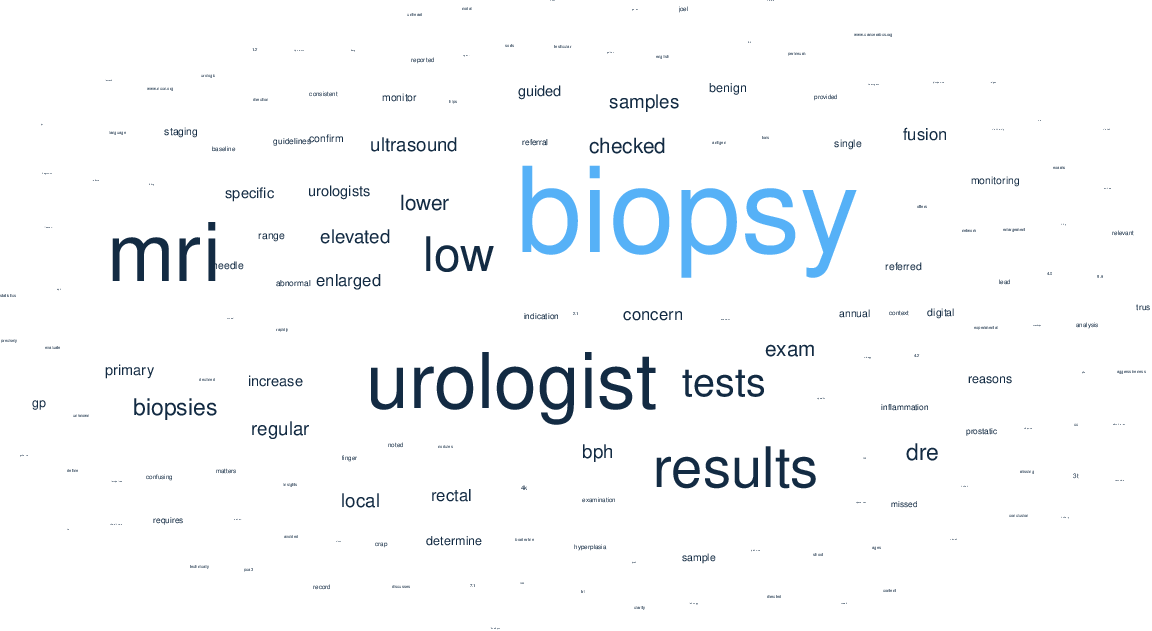}
        \caption{Prognostic staging topic}
        \label{fig:biopsy}
    \end{subfigure}
\hfil
    \begin{subfigure}{0.49\linewidth}
        \includegraphics[width=\linewidth]{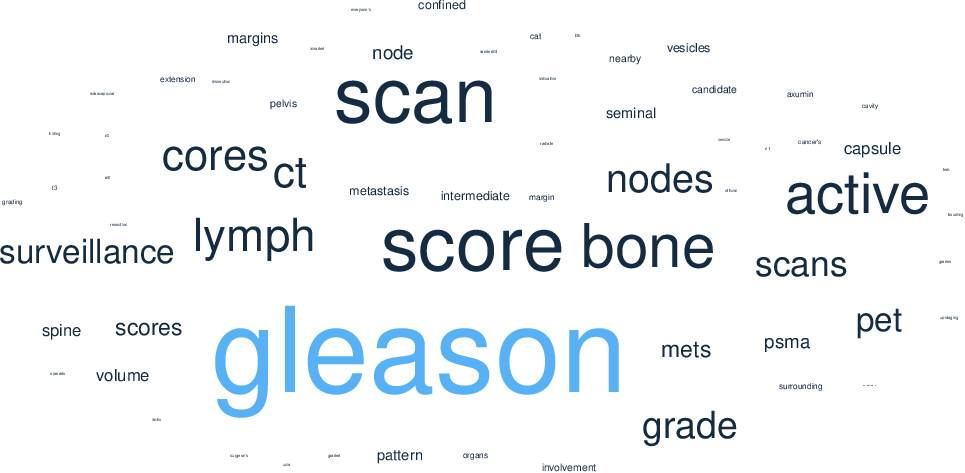}
        \caption{Prognostic scoring topic}
        \label{fig:gleason}    
        \end{subfigure}
\par\bigskip \par\bigskip
    \begin{subfigure}{0.49\linewidth}
        \includegraphics[width=\linewidth]{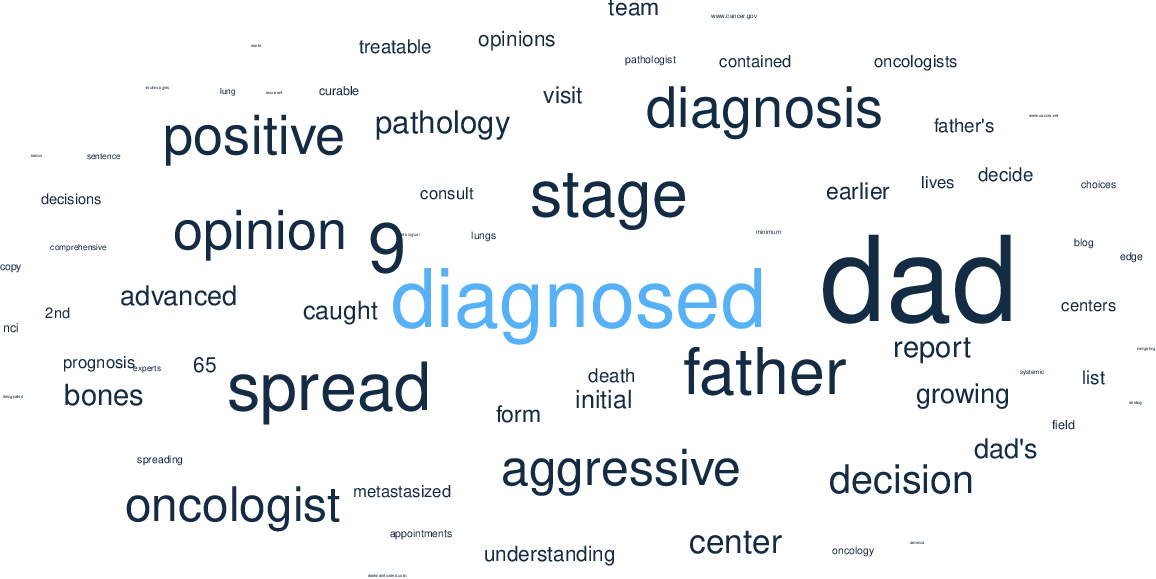}
    \caption{Diagnosis topic}
    \label{fig:diagnosed}  
    \end{subfigure}
\hfil
    \begin{subfigure}{0.49\linewidth}
        \includegraphics[width=\linewidth]{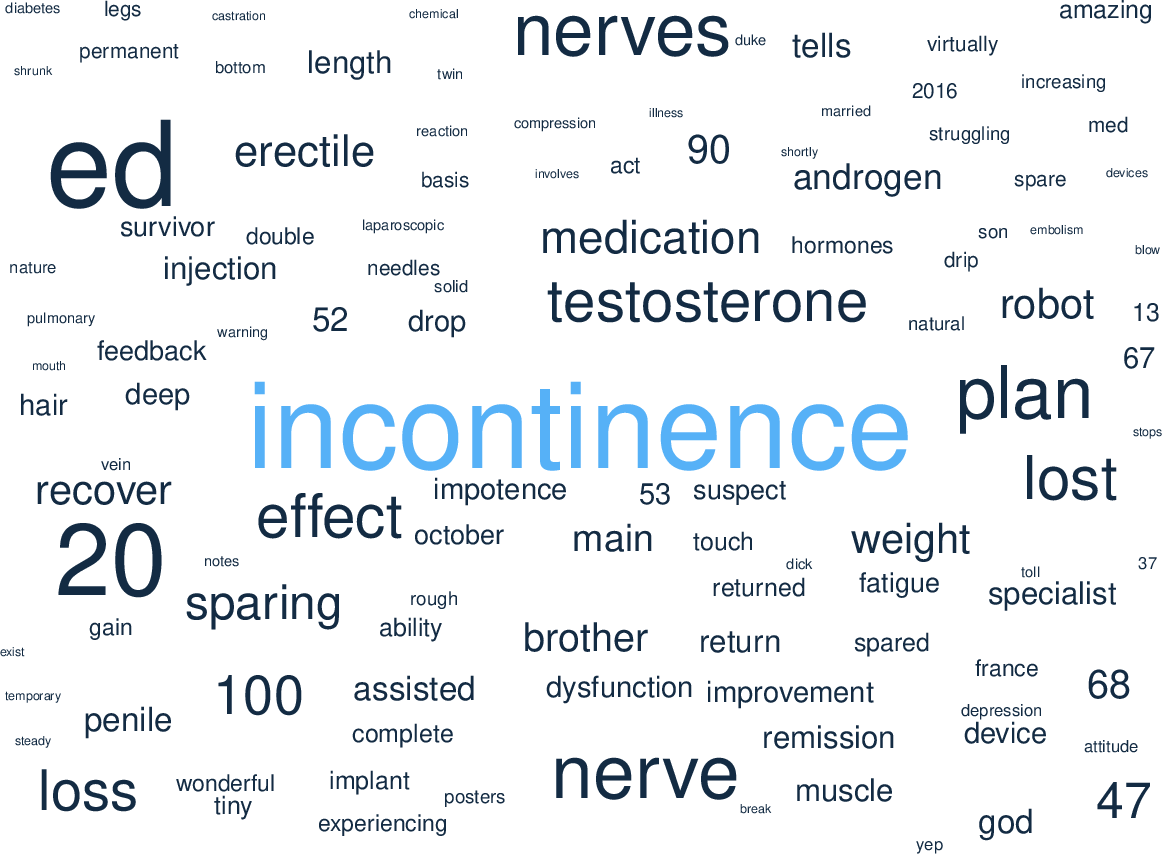}
    \caption{Prostate function topic}
    \label{fig:incontinence}  
    \end{subfigure}
\par\bigskip \par\bigskip 
    \begin{subfigure}{0.49\linewidth}
        \includegraphics[width=\linewidth]{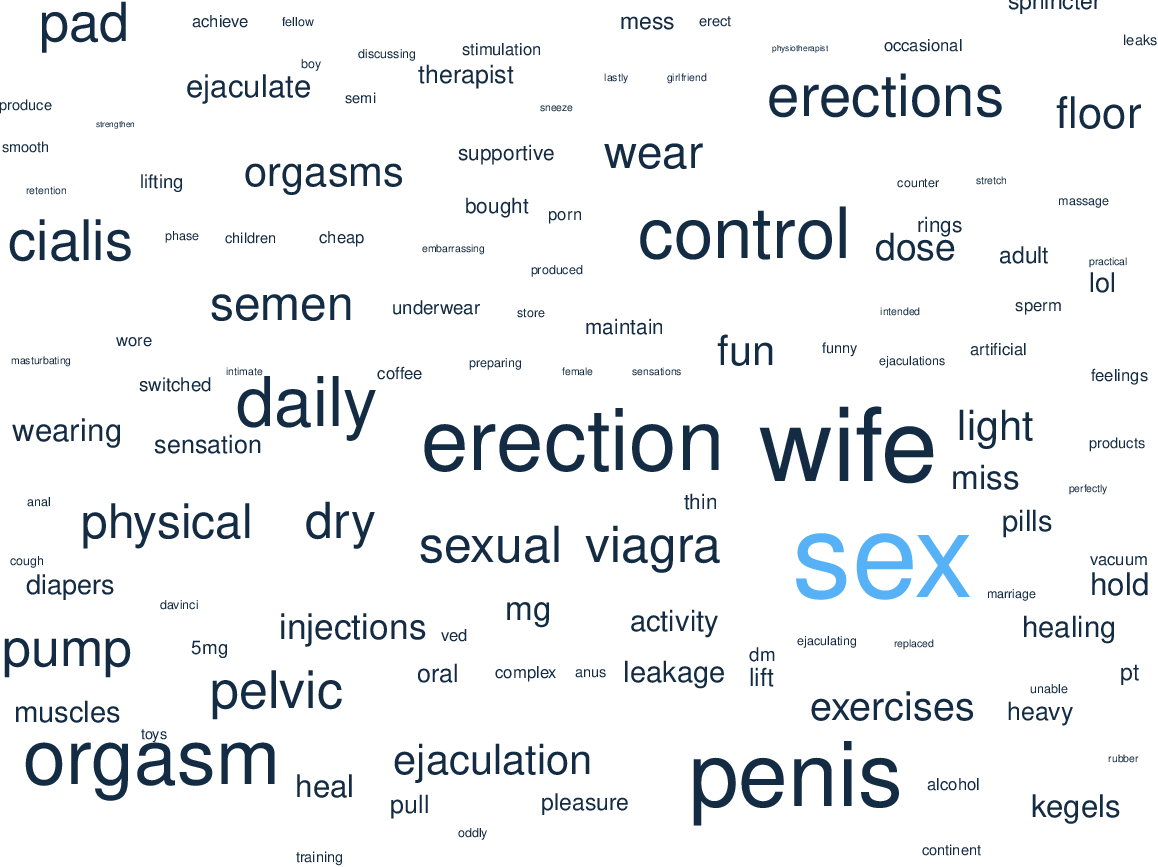}
    \caption{Sexual function topic}
    \label{fig:sex}  
    \end{subfigure}
\hfil
    \begin{subfigure}{0.49\linewidth}
        \includegraphics[width=\linewidth]{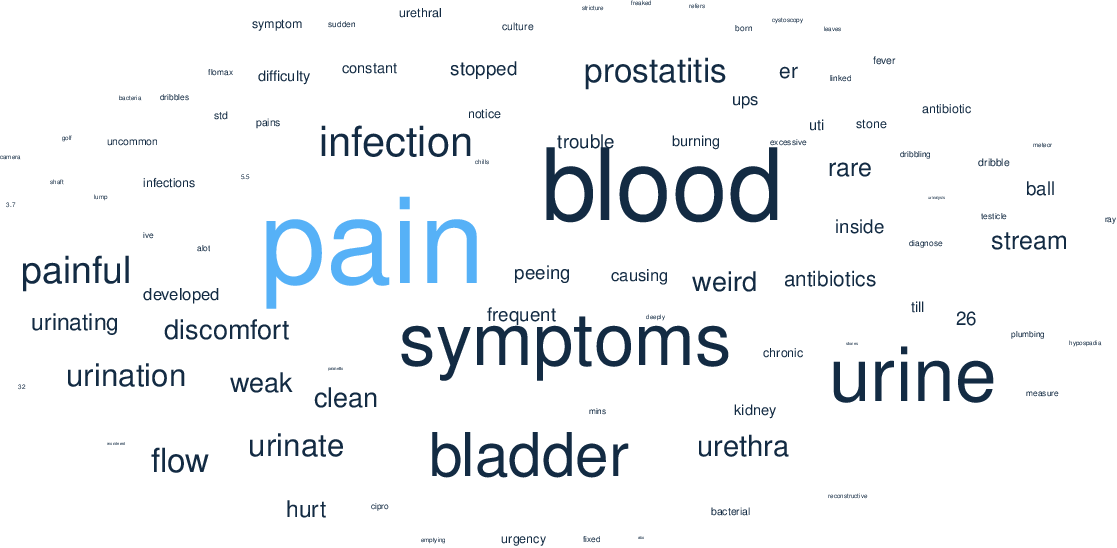}
    \caption{Prostatitis topic}
    \label{fig:pain}  
    \end{subfigure}
\caption{Six topics found through topic modelling of the/r/ProstateCancer discourse.}
    \label{fig:fig_all}
    \end{figure*}

Our topic model reveals a hierarchical structure with four layers. The first (most general) layer is a single topic further divided into two highly general topics in the second layer. The third layer contains twenty relatively general topics and the fourth layer 107 highly specific topics. We show six select topics of the twenty from the third layer in Figure \ref{fig:fig_all}.

% \begin{figure}[h!]  
%  \centering
%  \includegraphics[width=0.8\textwidth]{network.png}
%  \caption{Topic structure of prostate cancer data revealed in a radial tree.}
%   \label{fig:network}
% \end{figure}

% \begin{landscape}
% \begin{figure}[h!]  
%  \centering
%  \includegraphics[width=1.3\textwidth]{mds_full.eps}
%  \caption{}
%   \label{fig:mds_full}
% \end{figure}
% \end{landscape}

\subsubsection*{Prognostic Staging}

Prostate screening is often encouraged for men with an increased risk for prostate cancer. Digital rectal examinations (DRE) and prostate-specific antigen (PSA) tests are the most common screening tests. Abnormalities in these screenings, such as an enlarged prostate or elevated levels of PSA, indicate a potential risk for prostate cancer. If abnormalities are present, further diagnostic procedures are often referred. These include imaging procedures such as magnetic resonance imaging (MRI) and ultrasound, as well as prostate biopsy, wherein a tissue sample is removed from the prostate and analysed to determine if cancerous cells are present. These tests are often conducted by a urologist. We see that this specific aspect of the diagnostic pathway (except PSA testing) is captured in the `prognostic staging' topic, shown as a word cloud in Figure \ref{fig:biopsy}, with a density of 0.051, indicating the topic corresponds to 5.1\% of the discourse. 

% \begin{figure}[h!]  
%  \centering
%  \includegraphics[width=0.5\textwidth]{biopsy_1.eps}
%  \caption{}
%   \label{fig:biopsy}
% \end{figure}

\subsubsection*{Prognostic Scoring}

Prostate biopsy detects the presence of cancerous cells. If cancerous cells are detected, the biopsy tissue samples (cores) are analysed and given a Gleason score to determine how far the tissue deviates from healthy prostate tissue. If staging results are significant, further tests may be recommended to determine the spread of cancer beyond the prostate. These tests include bone scan, computed tomography (CT) scan, MRI, ultrasound, and positron emission tomography (PET) scan. Discourse relating to this aspect of the prostate cancer diagnostic process is evidenced in the `prognostic scoring' topic and captured in the word cloud in Figure \ref{fig:gleason}. The topic density for this topic is 0.028.

% \begin{figure}[h!]  
%  \centering
%  \includegraphics[width=0.5\textwidth]{gleason_1.eps}
%  \caption{}
%   \label{fig:gleason}
% \end{figure}

\subsubsection*{Diagnosis}

Discussion on the diagnosis of prostate cancer forms a topic, represented by the word cloud in Figure \ref{fig:diagnosed}. It has a density of 0.042, indicating that 4.2\% of the informative words in the discussion are drawn from this topic. The diagnosis discussion is characterised by the notable words `diagnosed' and `diagnosis'. This discussion appears to be often framed from the perspective of a child of a prostate cancer patient, with the paternal words `dad', and `father' being prominent words in the topic. The additional words in the topic: `spread', `aggressive', and `stage' indicate this topic corresponds to discussion of the stage that the cancer has reached. Words such as `bones', `lungs', and `metastasised' may indicate late-term cancer that has begun to spread to other sites in the body. As part of the diagnosis topic, we see some discussion related to the outlook and treatment methods available, noted by the words `prognosis', `curable', `treatable', `opinions', `decisions', and `decision'.

% \begin{figure}[h!]  
%  \centering
%  \includegraphics[width=0.5\textwidth]{diagnosed_1.eps}
%  \caption{}
%   \label{fig:diagnosed}
% \end{figure}

% \begin{figure}[h!]  
%  \centering
%  \includegraphics[width=0.8\textwidth]{surgery_1.eps}
%  \caption{}
%   \label{fig:surgery}
% \end{figure}

\subsubsection*{Prostate Function}

Figure \ref{fig:incontinence} evidences discussion on /r/ProstateCancer evolving into conversation related to sensitive topics, such as incontinence and erectile dysfunction (ED), two major side effects of the removal of the prostate in a procedure known as a radial prostatectomy (RP). This topic appears with a density of 0.050, indicating that it corresponds to 5\% of the discourse. The use of the words `nerve', `erectile', `dysfunction', `ed', `impotence', and `ability' can all be interpreted in the context of erectile dysfunction. There is a presence of the words `lost', `loss', and `shrunk', which appear to indicate that there is a worsening in the condition, and conversely, `returned', `return', `remission', `improvement', `increasing', `amazing', and `wonderful'. We see that discussion around penile shrinkage, a result of a loss of blood flow after prostatectomy, is captured in this topic through the former words, as well as `length'.

\subsubsection*{Sexual Intercourse}

% \begin{figure}[h!]  
%  \centering
%  \includegraphics[width=0.5\textwidth]{incontinence_1.eps}
%  \caption{}
%   \label{fig:incontinence}
% \end{figure}

Figure \ref{fig:sex} shows discussion related to sexual intercourse with a density of 0.051 and appears highly related to the topic in Figure \ref{fig:incontinence}. Whereas the topic in Figure \ref{fig:incontinence} appears to show discussion on sexual function and the lack thereof, the topic in Figure \ref{fig:sex} appears to go into more detail about the sexual experience of people with prostate cancer.

% \begin{figure}[h!]  
%  \centering
%  \includegraphics[width=0.5\textwidth]{sex_1.eps}
%  \caption{}
%   \label{fig:sex}
% \end{figure}

\subsubsection*{Prostatitis}

Figure \ref{fig:pain} illustrates discussion related to prostatitis symptoms, inflammation of the prostate, commonly due to bacterial infection. These symptoms may include urinary pain and burning, difficulty passing urine, `dribbling' (leaking urine), increased frequency of urination, blood in urine, generalised pain in the abdomen and genitals, and fever \cite{litwin2002review}. This discussion is observed through the use of the topic words `prostatitis', `urine', `urinating', `urination', `peeing', `flow', `stream', `difficulty', `weak', `dribble', `frequent', `pain', `painful', `discomfort', `hurt', `burning', and `fever'.

\subsubsection*{Other Topics}

Many other topics are included in the online supplementary materials as word clouds \cite{murray2022DAPMAV_github}. 
We found the topics primarily corresponded to five main categories, which we manually labelled as; `diagnosis', `treatment', `symptoms', `experience', and `general'. 

\subsection*{Topic Landscape}

\begin{figure*}
   \centering
   \includegraphics[width=\textwidth]{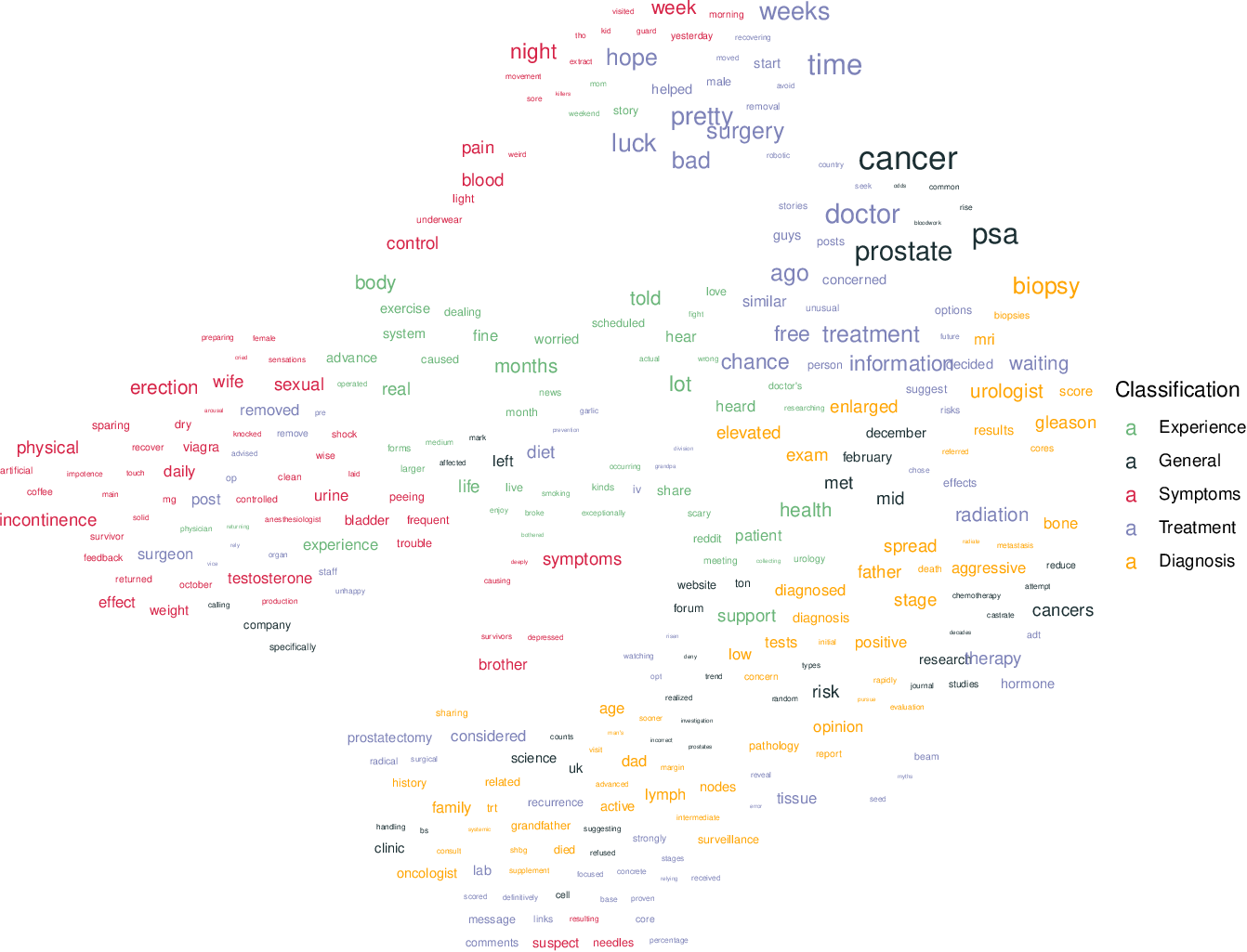}
   \caption{Topic landscape founds using UMAP with dissimilarity based on the negative co-occurrence rate between topics. The top three words from each topic at the lowest level in the topic hierarchy are shown and classified.}
    \label{fig:mds_part}
  \end{figure*}

The topic-topic interactions tell us the likelihood of topic co-occurrence, giving insights into the relationships between topics. As the network approach to topic modelling is built on a bipartite document-word network, within and between topic interactions are of probability zero. Edges in the network occur between words and documents, and hence the communities of words that form topics are connected only through their connections to communities of documents. We therefore use the two-step interaction matrix to retrieve topic-topic interactions, propagating edges from topics to documents, and back to topics. This interaction matrix gives a matrix of probabilities for topic co-occurrence, considering interactions within and between topics. We convert this to a dissimilarity measure by taking the negative co-occurrence rates. Using UMAP, we perform a dimension reduction from the topic-space to a two-dimensional space that seeks to preserve local distances between topics. This two-dimensional topic representation produces a map of the topics, seen in Figure \ref{fig:mds_part}. We note that here we are considering topics at the lowest (most specific) level in the topic hierarchy. We show the three most common words for each topic, scale the words' sizes to correspond to their relative frequencies, and colour according to the classifications made. 

\begin{table}
  \centering
  \caption{Table showing the most prevalent word in the topic (topic identifier) and the overall usage of the topic in the /r/ProstateCancer discourse.}
   \label{tab:topics}
 \begin{tabular}{lll}
 \toprule
 Topic & Figure & Usage density\\
 \midrule
 % surgery & \ref{fig:surgery} & 0.138\\
 % prostate & \ref{fig:prostate} & 0.071\\
 % prostatectomy & \ref{fig:prostatectomy} & 0.052\\
 Sexual function & \ref{fig:sex} & 0.051\\
 Prognostic staging & \ref{fig:biopsy} & 0.051\\
 Prostate functioning & \ref{fig:incontinence} & 0.050\\
 \addlinespace
 Diagnosis & \ref{fig:diagnosed} & 0.042\\
 Prostatitis & \ref{fig:pain} & 0.031\\
 Prognostic scoring & \ref{fig:gleason} & 0.028\\
 % radiation & \ref{fig:radiation} & 0.023\\
 \bottomrule
 \end{tabular} 
 \end{table}

  \subsection*{Topic Progressions and Emotional Arc}

\newcommand\ddfrac[2]{{\displaystyle\frac{\displaystyle #1}{\displaystyle #2}}}
 
 The average positional structure of topics may reveal the general trends that /r/ProstateCancer posts follow. We may gain insight into how these stories are told by capturing these trends. To find this positional structure, we find density estimates for the positions of topics throughout documents. First, we find normalised word positions in documents as the window of the document that the word occupies. For example, in a document consisting of two words: `Hello world.', the word `Hello' occupies the range $[0,0.5)$ and the word `word' occupies the range $[0.5, 1]$. The words can be replaced with their topics, and the positional densities taken as the relative frequency of the positional occupancy of the topic (relative to the topic's total use), say $\hat{t}$,
  $$p\left( x\right | \hat{t}) = \ddfrac{\sum_{d}\mathbf{1}_{\hat{t} = t^d_x}}
  {\sum_d \int_0^1\mathbf{1}_{\hat{t} = t^d_x}dx}
  ,$$
  where $t^d_x$ is the topic of document $d$ at position $x$, and $\mathbf{1}_{\hat{t} = t_x^d}$ is the indicator function,
  
  \begin{align*}
  \mathbf{1}_{\hat{t} = t_x^d}= \begin{cases}
  1, \quad \text{if } \hat{t} = t_x^d, \\
  0, \quad \text{otherwise.}
  \end{cases}
  \end{align*}
  
\begin{figure*}
   \centering
   \includegraphics[width=.7\textwidth]{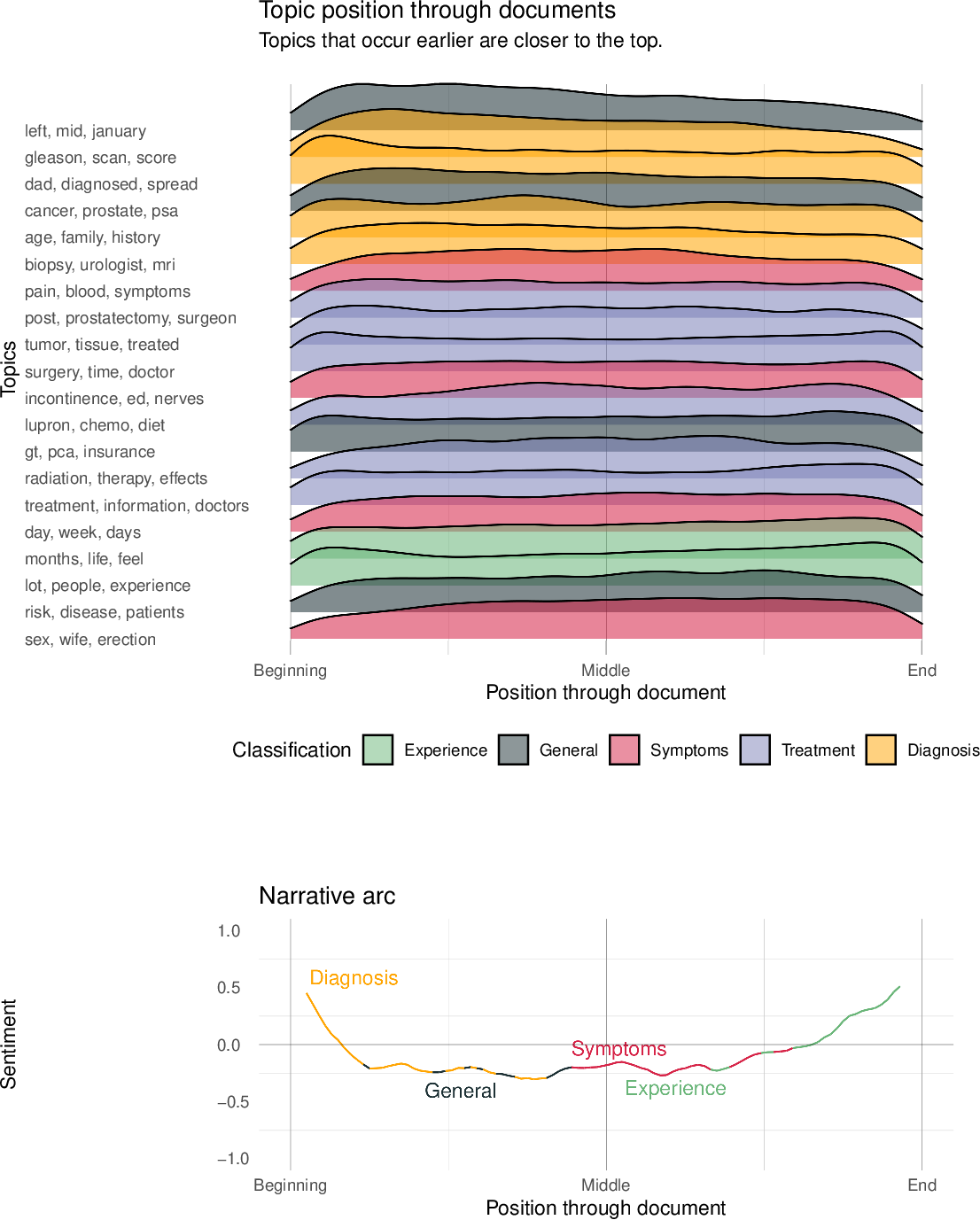}
   \caption{Distribution of topic discussion throughout the text.}
    \label{fig:location_1}
  \end{figure*}

Figure \ref{fig:location_1} shows stacked topic-position densities for each topic found at the second-lowest level of the topic hierarchy. The density stacks are arranged in ascending order of median position, i.e. topics discussed more frequently early (on average) are at the top of the stack. The lower half of Figure \ref{fig:location_1} shows the average emotional arc of the prostate cancer discourse by plotting the average sentiment against the normalised position through document \cite{reagan2016emotional}. The most dominant topic progression throughout the collective narrative is highlighted by segmenting the arc into the topics with the highest density at the respective positions.

%   \begin{figure}[h!]  
%   \centering
%   \includegraphics[width=\textwidth]{topic_location_1.eps}
%   \caption{Distribution of topic discussion throughout the text.}
%     \label{fig:location_1}
%   \end{figure}
  
%   \begin{figure}[h!]  
%   \centering
%   \includegraphics[width=\textwidth]{topic_location_density_lolipop_1.eps}
%   \caption{Distribution of topic discussion throughout the text.}
%     \label{fig:location_2}
%   \end{figure}

\section{Discussion}

\subsection*{DAPMAV Framework}
The DAPMAV framework introduced in this paper provides a structured approach for analysing patient-reported experiences found on social media. We follow the principles of this framework to elicit patient-reported experiences from prostate cancer discourse on the Reddit community /r/ProstateCancer.

The inductive reasoning underpinning this discussion -- in which real-world data is explored to generate ideas -- is established in grounded theory \cite{glaser2017discovery}. Uncovering structure in vast qualitative data sources reduces complexity. The uncovered structure reveals themes in the data. These themes can be codified to identify core concepts. Storylines can be built from the core concepts, which form the basis for model and hypothesis development. Tools in NLP allow us to uncover structure in online collections of patient-reported experience narratives. We demonstrate the utility of the DAPMAV framework to find and qualitatively analyse patient-reported experience discourse from social media discussions on prostate cancer. 

\subsection*{Topic Modelling}

Topic modelling distils the Reddit prostate cancer discourse into many themes. We codify these themes into five classifications. Two of the five classifications of discussion are related to clinical care. In particular, topics can be classified as `diagnosis' topics, representing topics corresponding to the discussion surrounding diagnosis and healthcare processes in the typical diagnostic pathway for prostate cancer, such as family history, screening, exams, biopsy, and diagnosis. The second of the clinical care topic classifications is the `treatment' classification for topics related to treatment pathways. The treatment pathway discourse is characterised by topics around active surveillance, prostatectomy, radiation therapy, androgen deprivation therapy, and others. The presence of these clinical care topics is important as it offers a new perspective on patient experience across diagnostic and disease treatment pathways. Two other classifications capture a more patient-centric arm of discussion, with communication of symptoms being one classification and patient experience being the other. Some examples of symptom-related topics are erectile dysfunction, sexual function, incontinence, pain, and anxiety. These are topics of a highly sensitive nature and reveal an intimate level of detail in their discussion that may be difficult to obtain in a survey, and likely much more difficult in a focus group. The patient experience classification encompasses topics such as quality of life, support, and questions. The final classification, `general', captures the remaining discussion and pertains to topics that may occur in many contexts, such as the mention of prostate cancer.

The prominence of paternal words in the diagnosis topic is a significant finding that sheds light on the involvement of family members in the prostate cancer journey. Terms such as `dad' and `father' indicate that family members of those impacted by prostate cancer actively seek community and support online. This insight serves as a reminder of the important role played by family in the care process. Involving and educating family members not only supports them in understanding and coping with the challenges their loved ones face, but also facilitates open and meaningful discussions within the family. By actively engaging family members, healthcare providers can create a supportive environment that encourages dialogue, shared decision-making, and holistic care. Recognising and addressing the needs of family members as stakeholders in the care process can lead to improved patient outcomes, enhanced emotional well-being for both patients and their families, and a more comprehensive and patient-centred approach to prostate cancer care.

One area of prostate cancer symptom discussion that was identified relates to sexual dysfunction, an unfortunate reality for many people who suffer from prostate cancer. Those who experience sexual dysfunction due to prostate cancer often indicate that the dysfunction negatively affects their quality of life. However, this suffering is frequently made in silence, as sexual dysfunction may be viewed by the patient as an embarrassing, emasculating illness with a high degree of perceived stigma \cite{fergus2002sexual,ettridge2018prostate}. The intimate and unconstrained discussion related to sexual dysfunction that can be found online offers an unfiltered window into the covert experiences faced by those who suffer from it. Topics seen in Figure \ref{fig:incontinence} and Figure \ref{fig:sex} capture the rich essence of these stories with expressive language and contribute to over $10\%$ of the total /r/ProstateCancer discourse. While sexual dysfunction as a result of prostate cancer is a well-explored area of research, it is important to note that in this case study, no costly exploratory study or focus group was needed to identify this area of concern. Instead, we detect this area of concern from public conversations that took place when and where the prostate cancer community pleased.

The in-depth and extensive discussions revolving around sexual dysfunction on the online platform /r/prostatecancer present a compelling case for the existence of a significant gap in patient satisfaction within traditional healthcare settings. These detailed conversations not only reflect the complexity of sexual concerns but also suggest that patients may not always feel adequately heard, understood, and supported when addressing such intimate issues in the healthcare system. The presence of profound social consequences, including feelings of embarrassment and the stigma associated with sexual dysfunction, further reinforces the motivation for individuals to seek reassurance and validation from online communities. Thus, the level of detail and engagement in these discussions serves as a poignant reminder of the unmet needs of patients, emphasising the pressing need for a more responsive and supportive healthcare environment that recognises and addresses the intricacies of sexual concerns.

As the word clouds shown in Figures \ref{fig:incontinence} and \ref{fig:sex} provide a rich overview of socially marginalising topics, taken directly from supportive communities that foster inclusion through shared experience, they have the potential to serve as powerful educational material that can lead to patient engagement. Healthcare providers and organisations can use these figures and results to develop informative brochures or web content that addresses sexual dysfunction to normalise this experience. By leveraging empirical experiences, patient-focused issues are represented by their own community, and thus highly relevant. The depth of experience, relatability, and rawness that these word clouds capture can equip patients with awareness, insights, and a sense of belonging that can spark discussions with their healthcare provider. In the same way, they could also be effective community-building assets. The importance of mitigating the negative impacts that prostate cancer can cause on one's quality of life cannot be understated; depression, anxiety, suicidal ideation, and suicidal mortality in patients with prostate cancer are higher than in the general population \cite{brunckhorst2021depression}.

Topic modelling allows batch processing of the prostate cancer corpus and reveals insightful characteristics of the nature of the corpus. We have demonstrated that well-researched areas of prostate cancer were readily detected without the need for costly surveys or focus groups. We gain an overview of each element of discussion and, importantly, draw comparisons between these elements and the real world. We qualitatively summarise the discourse by identifying topics with tangible aspects of healthcare. This approach in general, when applied to other areas allows for a qualitative exploration of patient views on healthcare, promoting the identification of areas of patient concern, as well as the potential to challenge our current views on them by comparing and contrasting our prior understanding with observations made.

\subsection*{Topic Landscape}

Analysis of specific topics provides specific insights related to the topics but does not capture the general theme of conversation. To summarise the entire discourse, we show a landscape of topics in Figure \ref{fig:mds_part}. This landscape forms a map of topics whose geographical regions correspond to clusters of topics that frequently occur together. These topics are represented by their three most prevalent words. 
Topics relating to patient experience are central to the map and shown in green. General topics are most prevalent in the top right of the figure and are coloured black. Topics relating to symptoms are shown in red and form a region on the left side of the figure that extends towards the top middle. The relative closeness of symptom and experience topics illustrates the frequent interactions between these types of topics. The treatment topic in purple is dispersed throughout the right half of the landscape and heavily interwoven with the diagnosis topic. This diagnosis topic is coloured yellow and extends from the bottom middle to the middle right. As UMAP produces this dimension reduction, global structure insights are traded off for greater visualisation of local structures. 

The emotional impact of the surgical response to prostate cancer on relationships is reflected in the topic landscape through the adjacency of words like `wife', `cried', `erection', `removed', `pre', and `post', to the far left of the plot. This juxtaposition of themes captures the emotional distress and the concerns regarding changes in sexual function following treatment. Interestingly, this cluster of words is situated at a noticeable distance from similar, yet more clinical terms, like `surgery' and `removal'. This spatial dichotomy reflects an underlying contrast in discourse; while `surgery' and `removal' may be indicative of more clinical and factual discussions, the terms `removed' and `remove' seem more raw and emotional, evocative of a more personal engagement with the loss associated with the surgical procedure. 

By capturing themes and their relationships in prostate cancer experiences, the topic landscape can also serve as a useful piece of educational material for prostate cancer patients and their support network. By helping patients contextualise their experience as a part of a broader collective experience within a community, the perceived stigma of their experience may be reduced. Additionally, this figure could help to train and remind healthcare practitioners of the web of concerns that prostate cancer patients face.

\subsection*{Topic Progressions and Emotional Arc}

The average sentiment of the prostate cancer discourse (Figure \ref{fig:location_1}) exhibits a fall-rise pattern. This pattern is one of six basic story arcs, commonly known as the `man-in-a-hole' plot. This plot follows a typical trajectory of tragedy, followed by resolution. The fall, or tragedy, is characterised by the diagnosis topic, leaving the sentiment to bottom out and stay negative for most of the discourse. This bottoming-out corresponds to peaks in a general discussion of prostate cancer, followed by a peak in the discussion of symptoms. During the rise, we see a peak in the use of topics related to patient experience, such as quality of life. This characterisation of the collective discourse into diagnosis, general discussion, and discussion of symptoms, followed by discussion of experience, is not entirely surprising. It follows a likely trajectory through healthcare. First, being diagnosed, then learning about the condition, experiencing symptoms as a result of the condition worsening or as side effects of treatment, and finally closing with some discussion on their experience. However, this pathway is somewhat incomplete, as discussion on treatment does not feature in this timeline. Note that this is due to the discussion on treatment being relatively evenly dispersed throughout the timeline. There are no strong peaks in the discussion of treatment across the timeline; hence, the peaks for other topic classifications are more prominent in comparison.

\subsection*{Limitations}
While the framework does not remove the need for technical proficiency in undertaking the analysis, it provides a structure that anyone can adopt. Methods such as network-based topic modelling can be computationally demanding when dealing with vast quantities of data. We propose that to overcome computational concerns, data can be sampled. Yet, with advances in computational power, these may become issues of the past. While social media gives everyone a voice, socioeconomic factors influence different demographics' presence online. This introduces biases into the results we find, and as such, it is important to be aware of these biases when drawing conclusions from these results. However, the DAPMAV framework enables us to serve minority demographics if they participate in specific online communities, as applying this framework to those communities would allow for a greater understanding of their unique issues.
Topics capture a theme of discussion using a family of words; however, the use of a word from a topic does not necessarily guarantee that the word is being used in the context of the topic. For example, the word `dad' appears in the diagnosis topic and hence has a strong association with the discussion surrounding diagnosis. However, the word `dad' may be freely used in another context. This potential misassociation does not mean that topics are not useful or informative; on the contrary, they are. Topics arrive from the compression of textual information and hence provide a useful simplifying representation of discussion that assists in forming an overview of the themes in a discussion and the words used in the themes \cite{gerlach2018network}. For an overview to be useful, it must simplify a concept. No simplification comes without \textit{some} loss of information, for there is no such thing as a free lunch. Instead, a balance must be struck between retaining enough information for insights to be drawn and reducing cumbersome complexity that detracts from understanding.

\subsection*{Future Research}
This paper has outlined several avenues for future research. Firstly, the DAPMAV framework has demonstrated effectiveness in analyzing discussions related to prostate cancer. There is potential in employing this framework for the analysis of discussions concerning other diseases. Other diseases can be assessed independently to capture their associated patient-reported experiences. Additionally, a comparative analysis can be conducted to evaluate common elements in these discussions. For example, we found that stigmatised topics such as sexual dysfunction are often prominent in prostate cancer discussions online. In relation to other illnesses or addictions, different stigmatised topics may emerge, such as mental health issues in chronic illnesses or societal stigmas associated with substance use disorders. A deeper understanding and comparison of these elements may provide insights into the motivations driving individuals to seek support.

Furthermore, social marginalisation has been recognised as a driving factor for the prominence of self-help groups \cite{davison2000talks}. While this finding is significant, there is an opportunity to delve deeper into the relationship between social marginalisation and discussions in online communities related to various diseases. By examining the prevalence of topics that are socially marginalising within different disease-related online communities, we can further our understanding of how social factors influence the seeking of support and community engagement. This, in turn, can inform strategies for better addressing the needs of patients and affected individuals.

Another potential application of the DAPMAV framework involves exploring geographically identified data to explore and compare patient experiences both within and across borders. This could shed light on regional variations in patient experiences and issues, and potentially unveil diverse cultural perspectives and healthcare system influences.

In general, the DAPMAV framework can be particularly valuable when applied to data from marginalised minority groups. This approach has the potential to capture and highlight the unique challenges and issues faced by these communities, which may be underrepresented in research. By bringing attention to these issues through data-driven insights, there is an opportunity to influence and shape policies in a way that is more inclusive and considerate of the diverse needs of all populations. Such efforts are critical for fostering healthcare environments that are equitable and tailored to the multifaceted experiences of patients across different socio-cultural backgrounds.

\section{Conclusion}

%In this paper, we motivated the need for a generalised reproducible framework for conducting social media analysis of healthcare discourse to elicit patient-reported experiences. To meet this need, we proposed the Design-Acquire-Process-Model-Analyse-Visualise (DAPMAV) framework as an overview of qualitative techniques to explore social media patient-reported experiences at scale. This patient-centered approach presents a framework complementary to surveying to explore areas of patient concern and experience. We demonstrated the utility of the framework in capturing patient-reported experiences in a case study on discourse in the Reddit prostate cancer community /r/ProstateCancer. We elucidated the complexity and interactions of themes within patient discourse and revealed the prevalence of detailed discussions concerning sensitive issues such as sexual dysfunction, as well as showing that family members also sought support in the online community. Our findings have implications not only for the enhancement of healthcare practitioners’ understanding, but also have strong merit in the education and empowerment of patients and their families; however, it is necessary to acknowledge the limitations related to demographic biases relating to social media data. Future research should investigate the application of the DAPMAV framework in diverse contexts, enabling an insightful view within and between areas of patient concern.

In this paper, the necessity for a generalised, reproducible framework to conduct social media analysis in healthcare discourse was motivated, with an objective of eliciting patient-reported experiences. To address this need, the Design-Acquire-Process-Model-Analyse-Visualise (DAPMAV) framework was proposed, encompassing an array of qualitative techniques to systematically explore patient-reported experiences on social media. This framework offers a complementary approach to traditional survey methods in a scalable patient-centred approach, affording a distinct exploration of patient concerns within community discussion. Through a case study analysing discourse in the Reddit prostate cancer community /r/ProstateCancer, this research demonstrates the utility of the DAPMAV framework in capturing patient-reported experiences. The analysis elucidated the depth, and interactions of themes within patient discourse, revealing not only detailed discussions on sensitive issues such as sexual dysfunction but also the engagement of family members seeking support within the online community. These findings, and figures that we produce have implications for enhancing healthcare practitioners' understanding of patient experience, but also have strong merit for fostering the education and empowerment of patients and their families. It is necessary, however, to recognise the limitations inherent in social media data, particularly with respect to demographic biases. Future research endeavours should focus on applying the DAPMAV framework in a variety of contexts, which could facilitate a more thorough understanding of patient experiences, and foster the development of healthcare policies and practices that are more aligned with patient needs and concerns.

% \begin{acks}
% None.
% \end{acks}

% \begin{dci}
% The authors declare that there is no conflict of interest.
% \end{dci}

% \begin{contribution}
% All authors contributed to the design of the framework. CM conducted the data collection and analysis with feedback from all other authors. CM wrote the first draft of the manuscript. All other authors reviewed and edited the manuscript. All authors approved the final version of the manuscript.
% \end{contribution}

% \begin{sm}
% Ethics approval was not required for this study.
% \end{sm}

% \begin{funding}
% The authors disclosed receipt of the following financial support for the research, authorship, and/or publication of this article: This work was supported by The University of Adelaide, LM is supported by the Australian Government through the Australian Research Council’s Discovery Projects funding scheme (project DP210103700).

% %[grant number yyy].
% %The author(s) received no financial support for the research, authorship, and/or publication of this article.
% \end{funding}

% \begin{guarantor}
% LM.
% \end{guarantor}
\bibliography{bibliography}

\appendix

\section{Supplementary Materials} \label{ap:sup}

Table \ref{tab:survey} shows the Australian Hospital Patient Experience Question Set. Table \ref{tab:key} shows the Google Scholar search terms from search keys seen in Table \ref{tab:fb_tw_re}.

\begin{table}
\caption{Key for Table \ref{tab:fb_tw_re} showing the exact Google Scholar search term for each Search Key. The term \texttt{[social media platform]} is a placeholder for the social media platform names; Facebook, Reddit, and Twitter.}
\label{tab:key}
\centering
\begin{tabular}[t]{>{\raggedright\arraybackslash}p{2cm}p{5.5cm}}
\toprule
Search Key & Google Scholar Search Term  \\
\midrule
- & \texttt{[social media platform]} \\[4em]

Patient Experience & \texttt{[social media platform] AND 
   `patient experience' OR `patient experiences' OR `patient reported experience' OR 
   `patient reported experiences'} \\[8em]
   
Natural Language Processing & \texttt{[social media platform] AND `natural language processing' OR `nlp'} \\[4em]

Patient Experience + Natural Language Processing & \texttt{[social media platform] AND 
   `patient experience' OR `patient experiences' OR `patient reported experience' OR 
   `patient reported experiences' AND `natural language processing' OR `nlp'} \\[4em]
\bottomrule
\end{tabular}
\end{table}

\begin{table*}
\caption{Australian Hospital Patient Experience Question Set (AHPEQS), a core set of satisfaction questions that hospitals can use to capture patient-reported experiences \cite{ACSQHC}.\label{tab:survey}}
\centering
\begin{tabular}[t]{>{\raggedright\arraybackslash}p{7cm}p{6cm}}
\toprule
Survey Questions & Response options\\
\midrule
My views and concerns were listened to & Always; Mostly; Sometimes; Rarely; Never; Didn't apply \\ \addlinespace
My individual needs were met (if answer always/mostly, skip to Q4) & Always; Mostly; Sometimes; Rarely; Never\\ \addlinespace
When a need could not be met, staff explained why & Always; Mostly; Sometimes; Rarely; Never\\ \addlinespace
I felt cared for & Always; Mostly; Sometimes; Rarely; Never\\ \addlinespace
I was involved as much as I wanted in making decisions about my treatment and care & Always; Mostly; Sometimes; Rarely; Never\\ \addlinespace
\addlinespace
I was kept informed as much as I wanted about my treatment and care & Always; Mostly; Sometimes; Rarely; Never\\ \addlinespace
As far as I could tell, the staff involved in my care communicated with each other about my treatment & Always; Mostly; Sometimes; Rarely; Never; Didn't apply\\ \addlinespace
I received pain relief that met my needs & Always; Mostly; Sometimes; Rarely; Never; Didn't apply\\ \addlinespace
When I was in the hospital, I felt confident in the safety of my treatment and care& Always; Mostly; Sometimes; Rarely; Never\\ \addlinespace
I experienced unexpected harm or distress as a result of my treatment or care ([if answer is no, skip to Q12) & Yes, physical harm; Yes, emotional distress; Yes, both; No\\ \addlinespace
\addlinespace
My harm or distress was discussed with me by staff & Yes; No; Not sure; Didn't want to discuss it\\ \addlinespace
Overall, the quality of the treatment and care I received was: & Very good; Good; Average; Poor; Very poor\\ \addlinespace
\bottomrule
\end{tabular}
\end{table*}

\end{document}